\documentclass[sigconf]{acmart}
\AtBeginDocument{%
  }

\copyrightyear{2025}
\acmYear{2025}
\setcopyright{cc}
\setcctype{by}
\acmConference[MM '25]{Proceedings of the 33rd ACM International Conference on Multimedia}{October 27--31, 2025}{Dublin, Ireland}
\acmBooktitle{Proceedings of the 33rd ACM International Conference on Multimedia (MM '25), October 27--31, 2025, Dublin, Ireland}\acmDOI{10.1145/3746027.3755120}
\acmISBN{979-8-4007-2035-2/2025/10}

\usepackage{multirow}
\usepackage{colortbl}
\usepackage{bbding}
\definecolor{lightpink}{rgb}{0.9, 0.95, 1.0}  
\usepackage{float}
\usepackage{subcaption}

\usepackage{caption} 
\usepackage{hyperref}

\usepackage{amsmath,amsfonts,bm}

\def\eqref#1{equation~\ref{#1}}

\def\1{\bm{1}}

\def\vc{{\bm{c}}}

\def\vv{{\bm{v}}}

\def\vx{{\bm{x}}}

\def\mC{{\bm{C}}}

\DeclareMathAlphabet{\mathsfit}{\encodingdefault}{\sfdefault}{m}{sl}
\SetMathAlphabet{\mathsfit}{bold}{\encodingdefault}{\sfdefault}{bx}{n}

\newcommand{\eg}{e.g.}
\usepackage{enumitem}
\begin{document}

\title{Knowledge Regularized Negative Feature Tuning of Vision-Language Models for Out-of-Distribution Detection}

\author{Wenjie Zhu}
\authornotemark[1]
\orcid{0000-0002-2791-3358}
\affiliation{%
  \institution{Hong Kong Polytechnic University}
  \city{Hong Kong}
  \country{China}
}
\affiliation{%
  \institution{Eastern Institute of Technology}
  \city{Ningbo}
  \country{China}
}
\email{22040319r@connect.polyu.hk}

\author{Yabin Zhang}
\affiliation{%
  \institution{Stanford University}
  \city{Stanford}
  \country{USA}}
\email{yabin@stanford.edu}
\authornote{Both authors contributed equally to this research.}

\author{Xin Jin}
\affiliation{%
  \institution{Ningbo Institute of Digital Twin, Eastern Institute of Technology}
  \city{Ningbo}
  \country{China}
}
\email{jinxin@eitech.edu.cn}

\author{Wenjun Zeng}
\authornotemark[2]
\affiliation{%
 \institution{Ningbo Institute of Digital Twin, Eastern Institute of Technology}
 \city{Ningbo}
 \country{China}}
 \email{wzeng-vp@eitech.edu.cn}

\author{Lei Zhang}
\affiliation{
  \institution{Hong Kong Polytechnic University}
  \city{Hong Kong}
  \country{China}
}
\email{cslzhang@comp.polyu.edu.hk}

\authornote{Corresponding author.}


\begin{abstract}

Out-of-distribution (OOD) detection is crucial for building reliable machine learning models. Although negative prompt tuning has enhanced the OOD detection capabilities of vision-language models, these tuned models often suffer from reduced generalization performance on unseen classes and styles. To address this challenge, we propose a novel method called Knowledge Regularized Negative Feature Tuning (KR-NFT), which integrates an innovative adaptation architecture termed Negative Feature Tuning (NFT) and a corresponding knowledge-regularization (KR) optimization strategy. Specifically, NFT applies distribution-aware transformations to pre-trained text features, effectively separating positive and negative features into distinct spaces. This separation maximizes the distinction between in-distribution (ID) and OOD images.
Additionally, we introduce image-conditional learnable factors through a lightweight meta-network, enabling dynamic adaptation to individual images and mitigating sensitivity to class and style shifts. Compared to traditional negative prompt tuning, NFT demonstrates superior efficiency and scalability. To optimize this adaptation architecture, the KR optimization strategy is designed to enhance the discrimination between ID and OOD sets while mitigating pre-trained knowledge forgetting. This enhances OOD detection performance on trained ID classes while simultaneously improving OOD detection on unseen ID datasets. Notably, when trained with few-shot samples from ImageNet dataset, KR-NFT not only improves ID classification accuracy and OOD detection but also significantly reduces the FPR95 by 5.44\% under an unexplored generalization setting with unseen ID categories. Codes can be found at \href{https://github.com/ZhuWenjie98/KRNFT}{https://github.com/ZhuWenjie98/KRNFT}.
\end{abstract}


\maketitle
\begin{figure*}[t]
    \centering
    \begin{subfigure}{0.32\textwidth}
        \includegraphics[width=\linewidth]{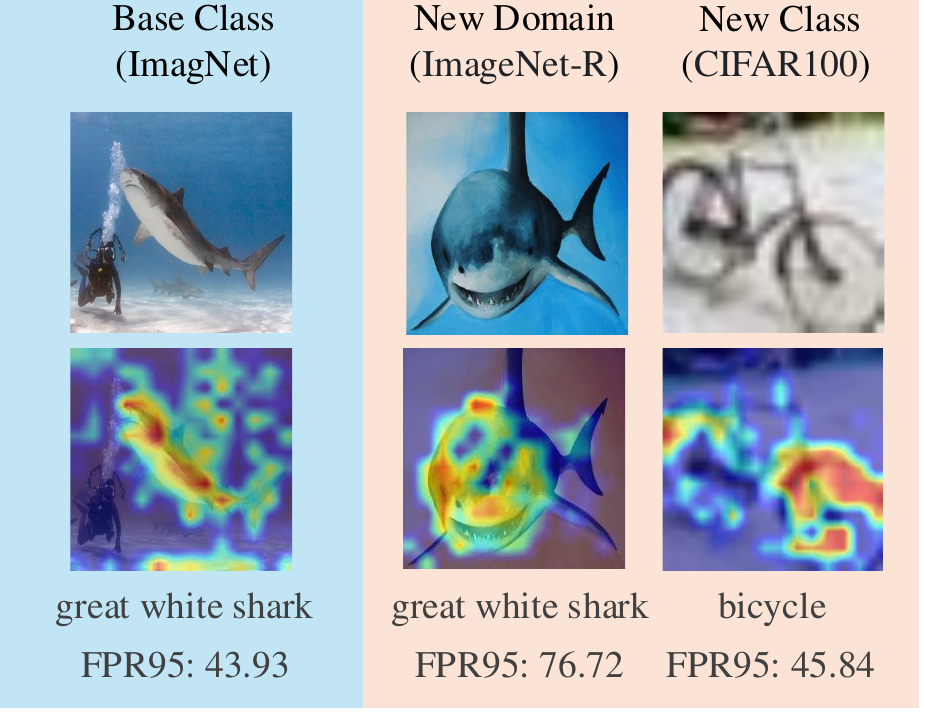}
        \caption{CLIP}
        \label{fig:clip}
    \end{subfigure}
    \begin{subfigure}{0.32\textwidth}
        \includegraphics[width=\linewidth]{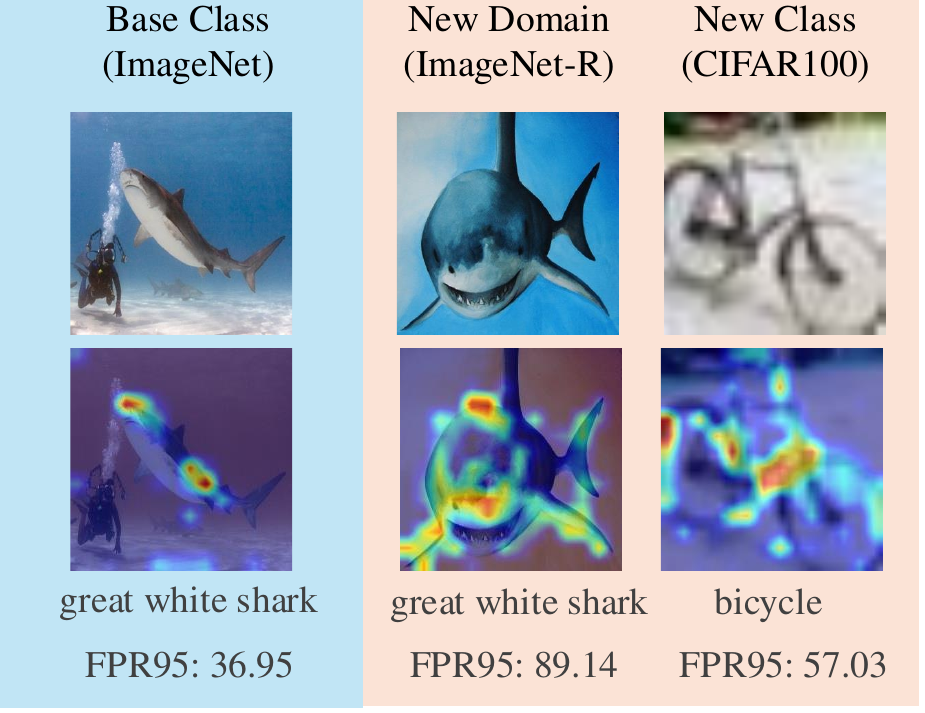}
        \caption{LoCoOp}
        \label{fig:locoop}
    \end{subfigure}
    \begin{subfigure}{0.32\textwidth}
        \includegraphics[width=\linewidth]{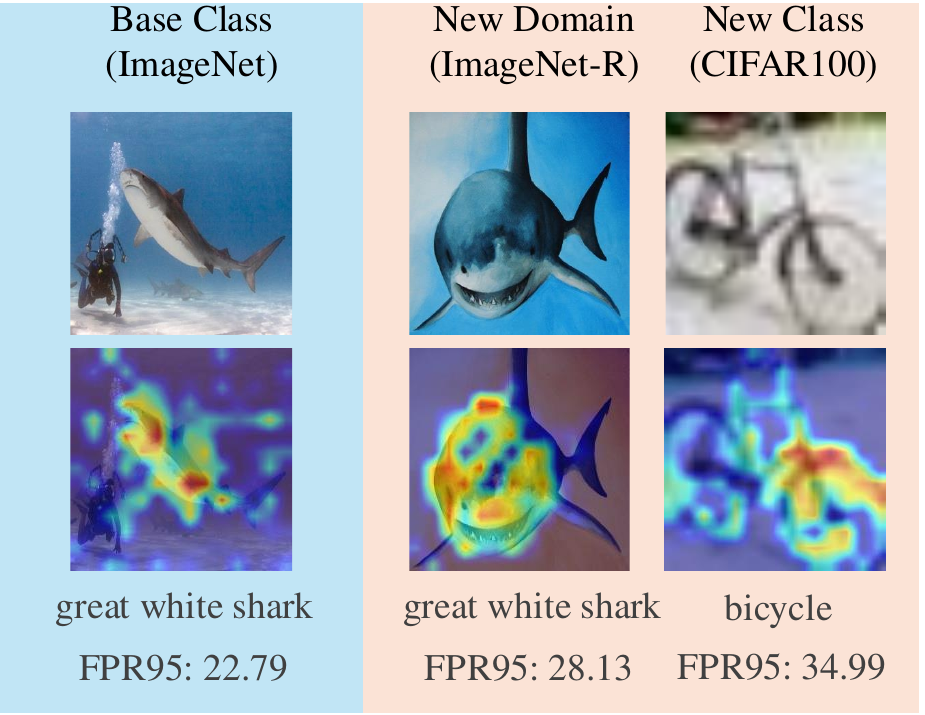}
        \caption{Ours(KR-NFT)}
        \label{fig:krft}
    \end{subfigure}
    \vspace{-0.3cm}
    \caption{GradCAM visualization of different methods on ID images. (a) In CLIP, the ID class shows high activation for both the foreground objects of ID images and the ID-irrelevant features. (b) In LoCoOp, although the activation of the ID class for ID-irrelevant features has decreased, there is also a reduction in the activation of the ID class for ID foreground objects when testing on unseen classes and styles. Consequently, the model's generalization performance for OOD detection has declined. (c) In our KR-NFT, the ID class shows a strong activation to ID foreground object when testing on unseen class and style, while exhibiting a low activation to ID-irrelevant features, demonstrating its strong comprehensive OOD detection capabilities.}
    \vspace{-0.3cm}
    \label{fig:motivation}
\end{figure*}

\vspace{-0.2cm}
\section{Introduction}
In real-world applications, machine learning models often encounter inputs from unknown classes, known as out-of-distribution (OOD) data.  Such OOD data can lead models to make erroneous predictions, posing significant safety risks, particularly in critical areas like autonomous driving \cite{bogdoll2022anomaly} and medical diagnostics \cite{anwar2018medical}. Consequently, the ability to detect these OOD samples is crucial in practice.

Traditional OOD detection methods in the image domain rely solely on visual information \cite{hendrycks2020pretrained, jin2022towards, hendrycks2016baseline, liang2017enhancing, liu2020energy, huang2021importance, wang2022vim, sun2022out, du2022unknown, tao2023non}. Recently, with the rise of vision-language models (VLMs) \cite{radford2021learning}, integrating textual information to improve OOD detection has gained increasing attention. Some initial studies have explored using pre-trained VLMs in a zero-shot manner \cite{ming2022delving, jiang2024negative,cao2024envisioning}, validating their strong capabilities. Recent efforts aim to further enhance their OOD detection abilities via model tuning, \eg, prompt tuning \cite{miyai2024locoop,nie2024out, bai2024id, li2024learning,zhang2024lapt}. By enhancing the relevance of ID classes to the foreground object of ID images and reducing the impact of ID-irrelevant features, these tuned models enhance the separation for ID and OOD images. Although these tuned models bring certain improvements to the training data, their generalization performance to unseen classes and styles has been significantly reduced, as shown in Fig. \ref{fig:motivation}. This indicates that current strategies fail to enhance the OOD detection capabilities of pre-trained VLMs comprehensively. 


We find that the reduced generalization performance of current tuning methods can be largely attributed to overfitting to training data and forgetting pre-trained knowledge. To address these issues, we propose a novel approach called Knowledge Regularized Negative Feature Tuning (KR-NFT).  Specifically, 
unlike prompt tuning methods, which abandon the pre-trained text features and build new ones, our method decouples pre-trained knowledge and new knowledge in the text feature space, enabling the model to have stronger generalization capabilities for various types of OOD detection. This novel feature-tuning strategy is characterized by three critical properties that enhance generalization in OOD detection. 
\textit{First}, the negative feature tuning directly introduces the distribution-aware learnable parameters on text features. By applying classification loss and OOD detection loss, the positive and negative text features can be optimized into different spaces. This creates a well-separated boundary to classify the ID and OOD images. \textit{Second}, the image-conditional transformation dynamically generates input-conditional factors for each image by a lightweight meta-network. Integrating instance-adaptive characteristics into the feature tuning can drive it to learn invariant representations \cite{derakhshani2023bayesian} and alleviate overfitting to class and style shifts. Furthermore, this image-conditional feature tuning shows high efficiency, as shown in Tab. \ref{tab:efficiency}. \textit{Third}, the knowledge regularization optimization strategy minimizes the discrepancy between the pre-trained text features and tuned text features, alleviating knowledge forgetting. Together, these properties significantly reduce the forgetting of pre-trained knowledge and overfitting to training data, thereby enhancing the OOD detection capabilities of VLMs comprehensively.

We conduct thorough analyses on the negative feature tuning structure and its corresponding knowledge regularization strategy. With extensive experiments, our KR-NFT not only achieves state-of-the-art performance on the trained ID classes but also presents strong generalization capabilities to unseen classes and styles. Especially, with models tuned on the ImageNet dataset, our method outperforms the closest competitor \cite{zhang2024lapt} by 5.44\% FPR95 on unseen classes and by 16.77\% FPR95 on unseen styles on average. Besides its outstanding performance, our method is characterized by its fast training and testing speeds, as analyzed in Tab. \ref{tab:efficiency}. Moreover, KR-NFT is highly scalable and can be seamlessly combined with and enhance other methods, as shown in Tab. \ref{tab:compatibility}.
We summarize our contribution as follows:
\begin{itemize}[left=0pt]
\vspace{-0.1cm}
\item    
We first identify a fundamental limitation of existing VLMs adaptation methods for OOD detection: while these tuned models bring improvements on the training data, they 
 often suffer from reduced generalization performance on unseen classes and styles.  

\item 
To tackle this issue, we propose a novel KR-NFT method by integrating an innovative NFT adaptation architecture and a corresponding KR optimization strategy. The NFT presents high efficiency and scalability, characterized by the distribution-aware transformation design and instance-adaptive image-conditional modulation. The KR balances the new task learning against pre-trained knowledge forgetting, enhancing the generalization capabilities of OOD detection on both trained and unseen ID datasets.

\item 
Validated with extensive experiments, our approach not only achieves new state-of-the-art results on the training data but also enhances the OOD detection capabilities of VLMs on unseen classes and styles. Especially, with models tuned on the ImageNet dataset, our method outperforms the closest competitor \cite{zhang2024lapt} by 5.44\% FPR95 on unseen classes and by 16.77\% FPR95 on unseen styles on average. 

    \vspace{-0.1cm}
\end{itemize}

\begin{figure*}[ht]
    \centering
    \includegraphics[width=0.98\textwidth]{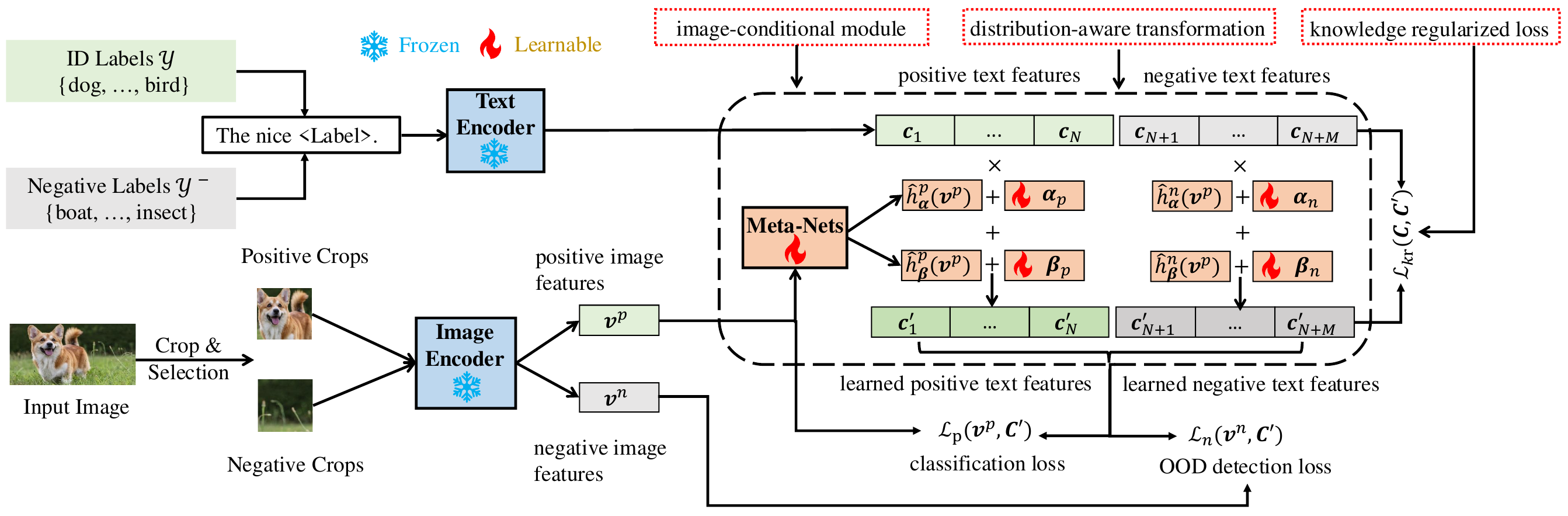}  
    \vspace{-0.2cm}
    \caption{The overall framework of our KR-NFT, where we introduce a knowledge regularized negative feature tuning with three critical properties, i.e., image-conditional module,  distribution-aware transformation, and knowledge regularized loss.
    These components are designed to enhance the generalization performance to unseen classes and styles in OOD detection.
    \vspace{-0.2cm}
    }
    \label{fig:framework}
\end{figure*}
\section{Related Work}

\textbf{Traditional OOD Detection.}
Traditional OOD detection methods can be roughly divided into three categories: classification-based \cite{hendrycks2016baseline,liang2017enhancing, lee2018simple, liu2020energy, sastry2020detecting, zhang2022out, sun2021react, dong2022neural, sun2022dice, lin2021mood, park2023nearest, jiang2023detecting, thulasidasan2019mixup,yun2019cutmix,wei2022mitigating,huang2021mos, hendrycks2018deep, fort2021exploring, yu2019unsupervised, papadopoulos2021outlier, ming2022poem}, density-based \cite{lee2018simple, zong2018deep, ren2019likelihood,abati2019latent, pidhorskyi2018generative, jiang2021revisiting, wang2022vim}, and distance-based \cite{lee2018simple, zaeemzadeh2021out, van2020uncertainty, ming2022exploit}.
Classification-based methods can be further divided into post-hoc  \cite{hendrycks2016baseline,liang2017enhancing, lee2018simple, liu2020energy, sastry2020detecting, zhang2022out,sun2021react, dong2022neural, sun2022dice, lin2021mood, park2023nearest, jiang2023detecting}, training-based \cite{thulasidasan2019mixup,yun2019cutmix,wei2022mitigating,huang2021mos}, and outlier exposure \cite{hendrycks2018deep, fort2021exploring, yu2019unsupervised, papadopoulos2021outlier, ming2022poem} approaches.  For example, Liang \emph{et al.} \cite{liang2017enhancing} enhanced the softmax score \cite{hendrycks2016baseline} with temperature scaling and small input perturbations.  
The confidence-estimating branch was designed in \cite{devries2018learning} for interpretable prediction. Hendrycks \emph{et al.} \cite{hendrycks2018deep} introduced real outliers to facilitate OOD detection. Density-based methods mainly rely on probabilistic models and predict low-density regions as OOD \cite{ren2019likelihood}. 
Conversely, distance-based methods differentiate OOD samples by calculating the distance between test data and ID, employing metrics such as the Mahalanobis distance \cite{lee2018simple} or nearest-neighbor distance \cite{sun2022out}.

\vspace{0.2cm}
\noindent\textbf{OOD Detection with VLMs.}
Enhancing OOD detection by incorporating language knowledge has gained increasing attention, which can be roughly divided into zero-shot \cite{ming2022delving, cao2024envisioning, jiang2024negative, zhang2024adaneg} and fine-tuning methods \cite{wang2023clipn, miyai2024locoop, bai2024id, li2024learning, zhang2024lapt}. MCM \cite{ming2022delving} pioneered the zero-shot setting by revisiting the softmax score \cite{hendrycks2016baseline} with pre-trained VLMs. Then, negative labels were introduced in \cite{jiang2024negative,cao2024envisioning} to further boost the performance. Recently, fine-tuning pre-trained VLMs to enhance OOD detection capabilities has been extensively studied \cite{wang2023clipn,nie2024out}, with prompt tuning emerging as the most popular technique \cite{miyai2024locoop,bai2024id,zhang2024lapt}. For instance, learning negative prompts corresponding to OOD samples has been explored in various studies \cite{wang2023clipn,nie2024out,bai2024id,zhang2024lapt}. However, we have observed that while these model-tuning methods achieve certain improvements in the training data, their generalization performance on unseen classes and styles is significantly reduced. This suggests that these approaches fail to enhance the OOD detection capabilities of VLMs comprehensively.

\vspace{0.2cm}
\noindent \textbf{CLIP Adaptation Methods.} 
The common CLIP Adaptation Methods include full fine-tuning, prompt tuning \cite{zhou2022learning, zhou2017places}, feature tuning\cite{gao2024clip, yu2023task, lian2022scaling, zhang2024dual}, and parameter tuning \cite{zanella2024low}. Among these, TaskRes \cite{yu2023task} and SSF \cite{lian2022scaling} are the methods most similar to our approach. TaskRes\cite{yu2023task} enhances classification performance on training data by adding a learnable residual vector to the text feature for each class, demonstrating its strengths in prior knowledge preservation and flexible task learning. SSF\cite{lian2022scaling} employs feature scaling and shifting parameters to boost a model's category recognition capabilities. However, the learning objectives of these methods are centered on classification, which limits their ability to effectively detect OOD samples. Consequently, nvestigating the utilization of these adaptation methods to bolster the robustness of CLIP against OOD samples emerges as a critical area.
\vspace{-0.4cm}
\section{Method}
\subsection{Preliminaries}
\textbf{OOD Detection.}
Define $\mathcal{X}$ as the image domain and $\mathcal{Y} = \{ y_1, \dots, y_N \}$ as the class label domain, with examples $\mathcal{Y} = \{ cat, dog, \dots, bird \}$ and $N$ denoting the total number of classes. We have $ \vx_{in} \in \mathcal{X}$ as the ID random variable and $\vx_{ood} \in \mathcal{X}$ as the OOD random variable, with their respective distributions $\mathcal{P}_{\vx_{in}}$ and $\mathcal{P}_{\vx_{ood}}$.
Typically, a test image $\vx$ is expected to follow the ID and to belong to one ID class, $\vx \in \mathcal{P}_{\vx_{in}}$ and $y \in \mathcal{Y}$ where $y$ is the label of $\vx$.
However, in practice, AI systems may encounter samples that do not match any known class, $\vx \in \mathcal{P}_{\vx_{ood}}$ and $y \notin \mathcal{Y}$, resulting in potential misclassifications and security issues \cite{scheirer2012toward,nguyen2015deep}.
To mitigate this, OOD detection distinguishes between ID and OOD samples using a scoring function $S$:
\begin{equation} \label{Equ:ood_score_function}
G_\gamma(\vx)= \begin{cases} \text{ID} & S(\vx) \geq \gamma \\ \text{OOD} & S(\vx) < \gamma \end{cases}
\end{equation}
where $G_{\gamma}$ is the OOD detector with threshold $\gamma$, determining ID if $S(\vx) \geq \gamma$.

\vspace{0.1cm}
\noindent\textbf{OOD detection with VLMs.}
Detecting OOD images by employing textual information has garnered increasing attention recently.
For a test image $\vx$ and the target label set $\mathcal{Y}$, we derive the image feature $\vv = f_{img}(\vx) \in \mathcal{R}^D$ and the textual features $\mC = f_{txt}(\rho(\mathcal{Y})) \in \mathcal{R}^{N\times D}$ using pre-trained encoders, where $D$ is the feature dimension. Here, $f_{img}(\cdot)$ and $f_{txt}(\cdot)$ represent the image and text encoders, respectively.
The function $\rho(\cdot)$ generates text prompts, typically formatted as `a photo of a $<$label$>$', where $<$label$>$ is replaced with specific class names such as `cat' or `dog'.
Both $\vv$ and $\mC$ are processed via $L_2$ normalization across the dimension $D$. 

To explore additional knowledge from a broader text space, NegLabel \cite{jiang2024negative} introduces negative labels $\mathcal{Y}^- = \{ y_{N+1}, \dots, y_{N+M} \}$ by identifying words that are far from the ID labels $\mathcal{Y}$ in extensive text corpora, leading to the following score function:
\begin{equation} \label{equ:neglabel_score}
    S_{NegLabel}(\vv) =  \frac{\sum_{i=1}^N {e^{\cos(\vv, \vc_i)}}}{\sum_{i=1}^N {e^{\cos(\vv, \vc_i)}} + \sum_{j=N+1}^{N+M} {e^{\cos(\vv, \vc_j)}}}.
\end{equation}

\vspace{0.1cm}
\noindent\textbf{Prompt Learning and Task Residual Learning.}
CoOp \cite{zhou2022learning} applies prompt learning to CLIP by adding some learnable context tokens in the input template. The prompt can be formulated as $\rho_p(y_i)=[\boldsymbol{V}]_1[\boldsymbol{V}]_2 \cdots[\boldsymbol{V}]_L[y_i]$ and each $[\boldsymbol{V}]_l$ is a vector with the same dimension as word embeddings of CLIP. To improve generalization performance, CoCoOp \cite{zhou2022conditional} introduces a lightweight Meta-Net 
 that generates each image an input-conditional token $h(x)$, then adds this token to each context token $\boldsymbol{V}_i(x) = \boldsymbol{V}_i(x) + h(x)$. Different from prompt learning, which introduces learnable context tokens on input, TaskRes \cite{yu2023task} introduces a portion of prior-independent parameters as a residual on the pre-trained classifier, by decoupling the old knowledge and the new target knowledge on the text features, TaskRes\cite{yu2023task} enhanced retention of existing knowledge and allows for more flexible investigation of task-specific knowledge. The new text features $\vc'$ can be formulated as: $\vc'= \vc + \mu\bm{\beta}$, $\mu$ is a scaling factor and $\bm{\beta}$ is a set of learnable parameters.
 
\subsection{Motivation}
Previous studies \cite{miyai2024locoop, bai2024id, li2024learning, zhang2022out} have explored applying prompt learning to enhance CLIP's capabilities on OOD detection. Though it can enhance the separation for ID and OOD images and bring certain improvements for OOD detection, these methods have a common issue: the tuned model's OOD detection performance on ID images from unseen classes and styles drops significantly, as illustrated in the Tab. \ref{tab:cross_datasets} and Tab. \ref{tab:domain_generalization}. This motivates us to the following research question: \\\\
{\centering\textit{How to enhance the generalization performance of VLMs in OOD detection?}} \\\\
To this end, we conduct experiments to investigate the correlation between the ID images and the positive text features. First, we find that CLIP has a strong alignment capability for ID images and ID labels, but it is easily affected by ID-irrelevant features, as shown in Fig. \ref{fig:clip}. Then, we observe that the drop in OOD detection for unseen ID classes in existing prompt learning methods is closely related to the reduction in the gradient weights of class activation mapping for ID foreground objects, suggesting a forgetting of CLIP's pre-trained knowledge, as shown in Fig. \ref{fig:locoop}. Therefore, a new framework is desired to balance the preservation of CLIP pre-trained knowledge and the learning of new tasks.

\vspace{0.1cm}
\subsection{Negative Feature Tuning.}
Existing works\cite{wang2023clipn,nie2024out,li2024learning,bai2024id} focus on designing negative prompts to empower CLIP with OOD detection capability. However, as illustrated in \cite{nie2024out}, employing a shared negative prompt fails to capture the diversity of negative features, and learning class-specific negative prompts will bring huge computational cost, as shown in the Tab. \ref{tab:efficiency}. To address these issues, we consider introducing learnable factors directly on negative text features. We first formulate the transformation process of the text feature $\mC$ in prompt tuning as:
\begin{equation} \label{equ:prompt_tuning}
\vc_i'=f_{t x t}(\rho_p(y_i))=T(\vc_i),
\end{equation}
where the prompt $\rho_p(y_i)=[\boldsymbol{V}]_1[\boldsymbol{V}]_2 \cdots[\boldsymbol{V}]_L[y_i]$ and each $[\boldsymbol{V}]_l$ is a vector with the same dimension. The tuned text feature $\vc_i'$ is a transformed output of the vanilla text feature $\vc_i$ generated by applying a general feature transformation function $T(\cdot)$. In other words, we could potentially achieve similar effects to prompt tuning by directly transforming pre-trained text features. To verify this, we experimented with various  implementations of transformation functions, including shifting by a constant, element-wise scaling/shifting, and MLP-based transformation, as analyzed in the Supplementary Material. We find that the simple element-wise scaling and shifting transformation achieves a good balance between capability and complexity. The learnable independent parameters are denoted as $\bm{\alpha}_{\text {neg }} = \left\{\omega_1^{\text{neg}}, \omega_2^{\text{neg}}, \ldots, \omega_n^{\text{neg}}\right\}$ and $\bm{\beta}_{\text{neg}} = \left\{b_1^{\text {neg}}, \omega_2^{\text{neg}}, \ldots, \omega_n^{\text{neg}}\right\}$, where $\bm{\alpha}_{\text {neg }} \in \mathcal{R}^D$ and $\bm{\beta}_{\text {neg }} \in \mathcal{R}^D$ are the element-wise scaling and shifting parameters, respectively. The negative feature tuning can ne denoted as:
\begin{equation} \label{equ:scale_shift}
\vc_i'= T(\vc_i) = L_2 (\bm{\alpha}_{\text{neg}} \vc_i + \bm{\beta}_{\text{neg}}),
\end{equation}
 $L_2 (\cdot)$ indicates the $L_2$ normalization along the feature dimension. 
\begin{figure}[ht]
    \centering
    \includegraphics[width=0.49\textwidth]{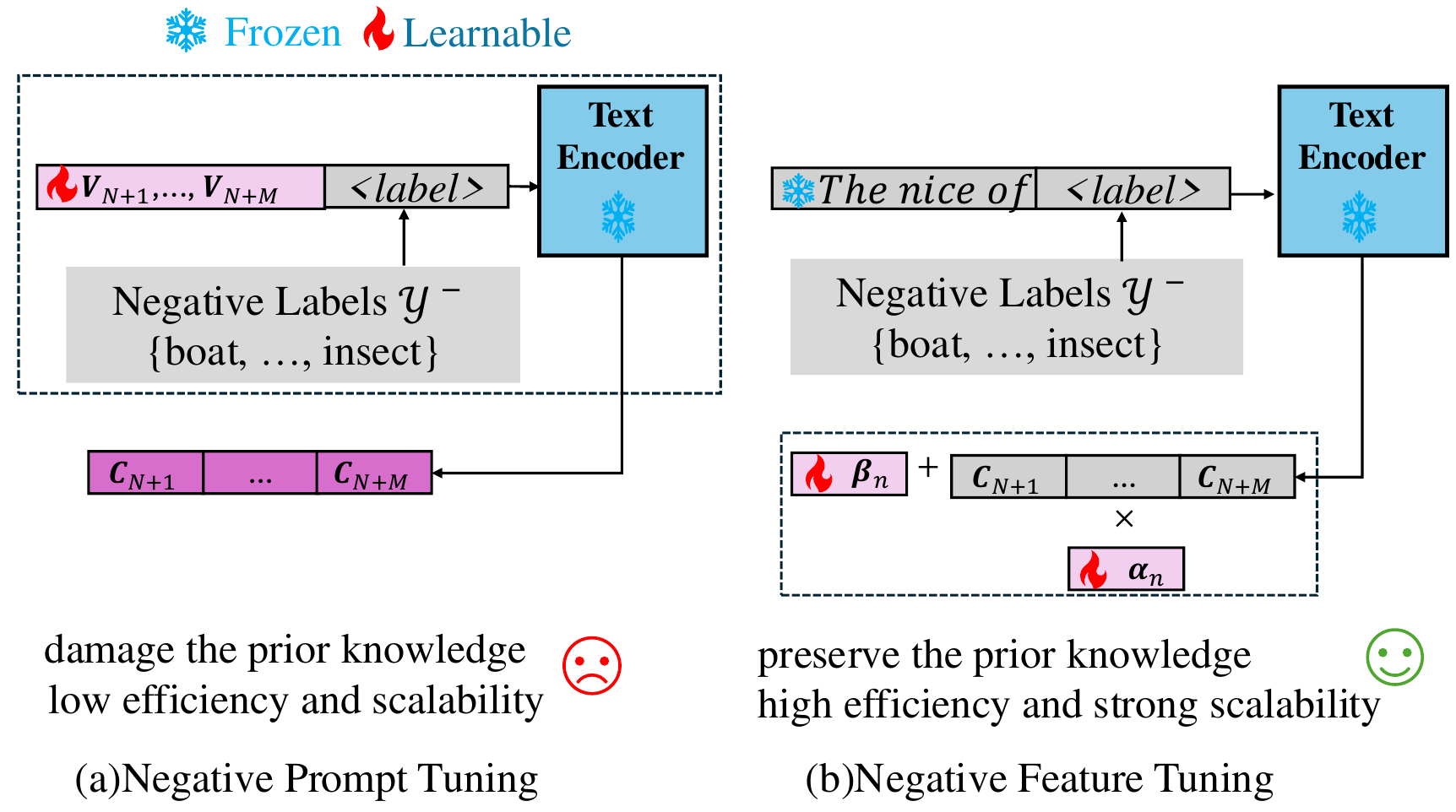}  
    \vspace{-0.4cm}
    \caption{Illustration of (a) Negative Prompt Tuning and (b) our Negative Feature Tuning. Our method introduces independent learnable factors directly on the pretrained features and does not require gradient backpropagation through the heavy text encoder. This demonstrates its advantage in preserving pre-trained knowledge and enhancing efficiency.
    }
    \vspace{-0.4cm}
    \label{fig:nft}
\end{figure}

\vspace{0.1cm}
\noindent\textbf{Image-Conditional Module.}
To enhance the generalization performance of learned prompts, conditioning the text prompts \cite{zhou2022conditional} on image features is a potential solution. We find that such conditional prompts indeed enhance the generalization performance; however, the testing speed is significantly reduced since it requires recalculating text features for each instance, as shown in Tab. \ref{tab:efficiency}. To balance the computational efficiency and generalization, we propose an image-conditional feature transformation module. Specifically, we can dynamically generate image conditional tokens for each test sample using neural networks, this can be formulated as:
\begin{equation} \label{equ:conditional_scale_shift}
\vc_i'= L_2 (h_{\bm{\alpha}}^{n}(\vv) \vc_i + h_{\bm{\beta}}^{n}(\vv)),
\end{equation}
The lightweight meta-net is denoted as $\widehat{h}_{\bm{\alpha}}^{n}(*)$  that takes image feature $\vv$ as input and outputs the residual parameters. Then, these conditional residual parameters will be added to the learnable  parameters $\bm{\alpha}$ to get a new image-conditional learnable parameters $h_{\bm{\alpha}}^{n}(\vv)$, which can formulated as $h_{\bm{\alpha}}^{n}(\vv) = \bm{\alpha} + \widehat{h}_{\bm{\alpha}}^{n}(\vv)$. And the  $h_{\bm{\beta}}(\vv)$ is similarly defined. The proposed image-conditional module can largely reduce the model's sensitivity to shifts in class and style, thereby enhancing its generalization capability, as validated in Tab \ref{tab:ablation}. Furthermore, it demonstrates strong advantages in computational efficiency,  as presented in Tab. \ref{tab:efficiency}.

\vspace{0.1cm}
\noindent\textbf{Distribution-aware Transformations.}
Unlike traditional classification tasks, which focus on a single ID, OOD detection tasks often introduce additional negative proxies to represent the negative distribution \cite{jiang2024negative,bai2024id}. Considering these varying distribution characteristics, we design a distribution-aware transformation functions for positive and negative text features, which can be formulated as:
\begin{align} \label{eq:distribution_aware_transformation}
    \vc_i' = \begin{cases} 
    L_2 (h_{\bm{\alpha}}^{p}(\vv) \vc_i + h_{\bm{\beta}}^{p}(\vv)), & \text{if } i \leq N \\
    L_2 (h_{\bm{\alpha}}^{n}(\vv) \vc_i + h_{\bm{\beta}}^{n}(\vv)), & \text{otherwise}
    \end{cases}
\end{align}
where $h_{*}^{p}$ and $h_{*}^{n}$ are the transformation functions for positive and negative text features, respectively. 
As analyzed in Figure \ref{fig:distributions}, such a distribution-aware setting significantly outperforms its distribution-agnostic counterpart.

\subsection{Knowledge Regularized Optimization.}
To enhance the separation capability between ID and OOD images through text features and avoid the text features overfitting on the training classes, we design a strategy named Knowledge Regularized Optimization. This strategy consists of three losses: traditional classification loss, ood detection loss, and pre-trained knowledge regularization loss. \\
\noindent\textbf{Outlier Samples Generation.}
We validate our method under the few-shot setting, where $S$ ID samples per class and a total of $K=SN$ ID instances are available for training, $\mathcal{D} = \{(\vx_1,y_1), ..., (\vx_K, y_K) \}$.
We follow \cite{bai2024id} to generate the negative training samples via image cropping and selection. Specifically, we apply multiple random cropping to each ID sample $\vx_k$ and get the cropping set $X_k^{crop} = \{ \vx_{k,1}^{crop}, ..., \vx_{k,P}^{crop}\}$, where $P$ is the number of random cropping. These croppings include positive samples that contain the target object, as well as negative samples do not contain any objects of interest (\eg, background only). We distinguish these two types of croppings by measuring their cosine similarity to the text feature of the corresponding label (\eg, `a photo of a $<$$y_k$$>$'). We define croppings with the highest and lowest cosine similarities as $X_k^{p} = \{ \vx_{k,1}^{p}, \ldots, \vx_{k,Q}^{p}\}$ and $X_k^{n} = \{ \vx_{k,1}^{n}, \ldots, \vx_{k,Q}^{n}\}$, where $Q$ is a hyperparameter that determines the number of selected croppings. Finally, we construct the training data as $\mathcal{D}_{p} = \{(\vx_{1,1}^{p}, y_1^{p}), (\vx_{1,2}^{p}, y_1^{p}), \ldots, \\
(\vx_{K,Q}^{p}, y_K)\}$ and $\mathcal{D}_{n} = \{(\vx_{1,1}^{n}, \vx_{1,2}^{n}, \ldots, \vx_{K,Q}^{n})\}$.
Our method is also compatible with other strategies that introduce negative training samples (\eg, exploring negative local features in LoCoOp \cite{miyai2024locoop}), as validated in the Supplementary Material.

\noindent\textbf{Traditional Classification Loss.}
Given the constructed model and the prepared training data, we optimize our method by maximizing the distinction between ID and OOD data. Specifically, we first employ the cross-entropy loss to enhance the model's ability to recognize ID classes:
\begin{align} \label{equ:id_loss}
    \mathcal{L}_{p}(\vx^{p}, y^{p}) =  
    -\log \frac{e^{\cos(\vv^{p}, \vc_{y^{p}}')}}{\sum_{i=1}^N {e^{\cos(\vv^{p}, \vc_i')}} + \sum_{j=N+1}^{N+M} {e^{\cos(\vv^{p}, \vc_j')}}},
\end{align}
where $(\vx^{p}, y^{p}) \in \mathcal{D}_{\text{p}}$, and $\vv^{p} = f_{img}(\vx^{p})$. 
Here, we reuse $y^{p}$ as the label indices of $\vx^{p}$ with a slight abuse of notation.
The scaling temperature is also omitted for readability.

\vspace{0.1cm}
\noindent\textbf{OOD Detection Loss.}
In addition to enhancing the recognition of ID classes, Eq. (\ref{equ:id_loss}) also pushes positive samples away from the negative labels. Similarly, we push negative samples away from the ID labels. Given that there is no one-to-one correspondence between negative samples and negative labels, we maximize the similarity between negative samples and the aggregate of negative labels. For optimization considerations, we adopt an equivalent objective by minimizing the similarity between negative samples and the aggregate of ID labels:
\begin{align} \label{equ:ood_loss}
    \mathcal{L}_{n}(\vx^{n}) = 
    \log \frac{\sum_{i=1}^N {e^{\cos(\vv^{n}, \vc_i')}}}{\sum_{i=1}^N {e^{\cos(\vv^{n}, \vc_i')}} + \sum_{j=N+1}^{N+M} {e^{\cos(\vv^{n}, \vc_j')}}},
\end{align}
where $\vx^{n} \in \mathcal{D}_{\text{n}}$, and $\vv^{n} = f_{img}(\vx^{n})$.

\vspace{0.1cm}
\noindent\textbf{Knowledge Regularization Loss.}
Although the adopted simple transformation function significantly preserves pre-trained knowledge, its implicitly structured design leaves room for further improvement. To further enhance the retention of pre-trained knowledge, we are inspired by the knowledge distillation framework \cite{hinton2015distilling} to introduce an explicit consistency constraint.
Specifically, we implement an objective that ensures the transformed text features remain consistent with the original ones in their outputs. 
As shown in Fig. \ref{fig:kp}, we investigated different locations for applying this consistency regularization, including on features, logits, or predicted probabilities, to determine the most effective approach.
The best performance is achieved by maximizing the cosine similarity between pre-trained text features $\vc_i$ and the learned $\vc_i'$:

\begin{equation}\label{eq:consistency_regularization}
\mathcal{L}_{\text {kr}}=\frac{1}{N+M} \sum_{i=1}^{N+M} (1 - \vc_i\vc_i').
\end{equation}

This consistency regularization objective can be understood as applying an additional constraint on the adopted transformation function. This constraint explicitly ensures that the transformations applied are subtle, in line with residual learning principles \cite{he2016deep}.  By limiting these transformations to minimal changes, this approach effectively preserve pre-trained knowledge, achieving a balance between acquiring new task capabilities and preserving the integrity of pre-trained knowledge. 

\begin{table*}[t] 
\small
\centering
\begin{tabular}{lcccccccc|cc}
\toprule
\multicolumn{11}{c}{OOD datasets}  \\
\multicolumn{1}{c}{{Methods}} & \multicolumn{2}{c}{INaturalist} & \multicolumn{2}{c}{SUN} & \multicolumn{2}{c}{Places} & \multicolumn{2}{c}{Textures} & \multicolumn{2}{c}{Average} \\ \cline{2-3} \cline{4-5} \cline{6-7} \cline{8-9} \cline{10-11}
 & \fontsize{8}{12}\selectfont AUROC$\uparrow$ & \fontsize{8}{12}\selectfont FPR95$\downarrow$& \fontsize{8}{12}\selectfont AUROC$\uparrow$ & \fontsize{8}{12}\selectfont FPR95$\downarrow$& \fontsize{8}{12}\selectfont AUROC$\uparrow$ & \fontsize{8}{12}\selectfont FPR95$\downarrow$& \fontsize{8}{12}\selectfont AUROC$\uparrow$ & \fontsize{8}{12}\selectfont FPR95$\downarrow$ & \fontsize{8}{12}\selectfont AUROC$\uparrow$ & \fontsize{8}{12}\selectfont FPR95$\downarrow$  \\
 \midrule
  \multicolumn{11}{c}{\textbf{Zero-shot methods)}} \\
MCM \cite{ming2022delving} & 94.59 & 32.20 & 92.25 & 38.80 & 90.31 & 46.20 & 86.12 & 58.50 & 90.82 & 43.93 \\
EOE \cite{cao2024envisioning} & 97.52 & 12.29 & 95.73 & 20.40 &  92.95 & 30.16 & 85.64 & 57.63 & 92.96 & 30.09 \\
NegLabel \cite{jiang2024negative} & 99.49 & 1.91 & 95.49 & 20.53 & 91.64 & 35.59 & 90.22 & 43.56 & 94.21 & 25.40 \\
\midrule
  \multicolumn{11}{c}{\textbf{Tuning-based methods}} \\ 
MSP \cite{hendrycks2016baseline} & 87.44 & 58.36 & 79.73 & 73.72 & 79.67 & 74.41 & 79.69 & 71.93 & 81.63 & 69.61   \\
ZOC \cite{esmaeilpour2022zero} & 86.09 & 87.30 & 81.20 & 81.51 & 83.39 & 73.06 & 76.46 & 98.90 & 81.79 & 85.19 \\
CLIPN \cite{wang2023clipn} & 95.27 & 23.94 & 93.93 & 26.17 & 92.28 & 33.45 & 90.93 & 40.83 & 93.10 & 31.10 \\
LSN \cite{nie2024out} & 95.83 & 21.56 & 94.35 & 26.32 & 91.25 & 34.48 & 90.42 & 38.54 & 92.26 & 30.22 \\
LoCoOp \cite{miyai2024locoop} & 93.93 & 29.45 & 90.32 & 41.13 & 90.54 & 44.15 & 93.24 & 33.06 & 92.01 & 36.95 \\
ID-like \cite{bai2024id} & 98.19 & 8.98 & 91.64 & 42.03 & 90.57 & 44.00 & \textbf{94.32} & \textbf{25.27} & 93.68 & 30.07 \\
NegPrompt \cite{li2024learning} & 90.49 & 37.79 & 92.25 & 32.11 & 91.16 & 35.52 & 88.38 & 43.93 & 90.57 & 37.34 \\
SCT \cite{li2024learning} & 95.86 & 13.94 & 95.33 & 20.55 & 92.24 & 29.86 & 89.06 & 41.51 & 93.27 & 26.47 \\
LAPT \cite{zhang2024lapt} & 99.63 & 1.16 & 96.01 & 19.12 & 92.01 & 33.01 & 91.06 & 40.32 & 94.68 & 23.40 \\
\rowcolor{lightpink} 
\textbf{KR-NFT (Ours)} & 99.62 & \textbf{0.82} & 96.15 & 17.83 & 92.64 & 36.12 & 91.96 & 36.38 & 95.09 & 22.79 \\
\rowcolor{lightpink}  \textbf{KR-NFT} ($\lambda_2$=0) & \textbf{99.67} & 1.33 & \textbf{96.28} & \textbf{17.46} & \textbf{93.68} & \textbf{28.17} & 93.26 & 29.34 & \textbf{95.82} & \textbf{19.08} \\
\bottomrule
\end{tabular}
\caption{OOD detection results for ID of ImageNet-1k and four OOD datasets using a VITB/16 encoder.
} \label{tab:four_ood_datasets}
\vspace{-0.8cm}
\end{table*}

\vspace{0.1cm}
\noindent\textbf{Learning Objectives.}
We construct our model based on NegLabel \cite{jiang2024negative}, which identifies OOD samples by comparing the similarity of image features to ID labels and mined negative labels, as shown in Eq. (\ref{equ:neglabel_score}). 
The overall learning objective is formulated as follows:
\begin{equation} \label{equ:final_loss}
\mathcal{L}=\mathcal{L}_{p}+ \lambda_1 \mathcal{L}_{n} +\lambda_2 \mathcal{L}_{kr},
\end{equation}
where $\lambda_1$ and $\lambda_2$ are balancing parameters. The overall framework is illustrated in Fig. \ref{fig:framework}.

\subsection{Inference.}
Existing fine-tuning methods for VLMs typically conduct evaluation on the training classes only \cite{zhang2024lapt,bai2024id,miyai2024locoop}. However, we found that these tuned models exhibit reduced generalization performance on unseen classes, as shown in Fig. \ref{fig:motivation}. This suggests that current tuning methods tend to overfit the training data and do not comprehensively enhance the VLMs' OOD detection capabilities. 
To comprehensively evaluate the OOD detection ability of tuned VLMs, we conduct inference not only on the trained ID classes but also directly evaluate the model, which is tuned on Base dataset, on unseen New classes.
Specifically, given a VLM typically tuned with samples $ \vx_{in} $ from the base ID distribution $\mathcal{P}_{\vx_{in}}$, we first assess its OOD detection performance using samples $ \vx_{in}/\vx_{ood} $ from $\mathcal{P}_{\vx_{in}}/\mathcal{P}_{\vx_{ood}}$, following the common pipeline \cite{huang2021importance,zhang2024lapt}. Additionally, we evaluate the same model with samples $\vx_{in}^{new}/\vx_{ood}^{new}$ from unseen class distributions $\mathcal{P}_{\vx_{in}^{new}}/\mathcal{P}_{\vx_{ood}^{new}}$, where the label space $\mathcal{Y}^{new}$ of $\mathcal{P}_{\vx_{in}^{new}}$ differs from the Base classes $ \mathcal{Y} $.
In addition to examining new distributions with class shifts, we also investigated another new distribution shift featured by image styles, which retains the same label space \( \mathcal{Y} \) as the Base classes and presents new image styles. We follow NegLabel to calculate the score in the testing stage, where we replace the pre-trained text feature $\vc_i$ with the learned $\vc_i'$ in Eq. (\ref{equ:neglabel_score}).

\begin{table*}[t] 
\small
\centering
\begin{tabular}{lcccccc|cc}
\toprule
\multirow{3}{*}{Methods}  & \multicolumn{8}{c}{Unseen New Classes}  \\
 & \multicolumn{2}{c}{CIFAR10} & \multicolumn{2}{c}{CIFAR100} & \multicolumn{2}{c|}{Fine-grained} & \multicolumn{2}{c}{Average} \\ 
  \midrule
 & \fontsize{8}{12}\selectfont AUROC$\uparrow$ & \fontsize{8}{12}\selectfont FPR95$\downarrow$& \fontsize{8}{12}\selectfont AUROC$\uparrow$ & \fontsize{8}{12}\selectfont FPR95$\downarrow$& \fontsize{8}{12}\selectfont AUROC$\uparrow$ & \fontsize{8}{12}\selectfont FPR95$\downarrow$ & \fontsize{8}{12}\selectfont AUROC$\uparrow$ & \fontsize{8}{12}\selectfont FPR95$\downarrow$  \\
 \midrule
  \multicolumn{9}{c}{\textbf{Zero-shot methods}} \\
  MCM \cite{ming2022delving} & 96.27 & 14.28 & 79.92 & 54.54 & 78.21 & 68.72 & 84.80 & 45.84 \\
  NegLabel \cite{jiang2024negative} & 96.69 & 12.15 & 80.21 & 54.05 & 89.99 & 41.28 & 88.96 & 35.82 \\
  \midrule
  \multicolumn{9}{c}{\textbf{Tuning-based methods}} \\ 
  LoCoOp \cite{miyai2024locoop} & 93.61 & 26.36 & 73.53 & 57.03 & 81.32 & 70.24 & 82.82 & 52.21 \\
  ID-like \cite{bai2024id} & 85.80 & 39.98 & 69.54 & 67.11 & 61.10 & 82.20 & 72.15 & 63.09 \\
  NegPrompt \cite{li2024learning} & 92.79 & 29.21 & 74.82 & 63.44 & 78.83 & 68.32 & 82.14 & 54.66 \\ 
  SCT \cite{yu2024self} & 95.17 & 18.28 & 77.55 & 62.99 & 83.11 & 65.29 & 85.28 & 48.85 \\ 
  LAPT \cite{zhang2024lapt} & 95.21 & 17.26 & 78.94 & 57.11 & 89.23 & 46.91 & 87.39 & 40.43 \\
\rowcolor{lightpink}  \textbf{KR-NFT (Ours)} & \textbf{96.82} & \textbf{11.41} & \textbf{80.95} & \textbf{53.24} & \textbf{89.99} & \textbf{40.33} & \textbf{89.25} & \textbf{34.99} \\
\bottomrule
\end{tabular}
   \caption{OOD detection results on unseen classes, where results of \cite{miyai2024locoop,bai2024id,li2024learning,zhang2024lapt} and ours are reported with models tuned on the ImageNet dataset. Pre-trained CLIP model is utilized in \cite{ming2022delving,jiang2024negative}. Detailed results are provided in the Supplementary Material.
    }
    \vspace{-0.6cm}
    \label{tab:cross_datasets}
\end{table*}

\vspace{-0.3cm}
\section{Experiments}
\subsection{Experiments Setup}
\noindent\textbf{Datasets and Benchmarks.} 
Following \cite{huang2021importance}, We select ImageNet-1K \cite{deng2009imagenet} as the ID dataset and use iNaturalist \cite{van2018inaturalist}, SUN \cite{xiao2010sun}, Places \cite{zhou2017places}, and Textures \cite{cimpoi2014describing} as OOD test datasets. 
To evaluate the generalization capability to unseen classes, we directly apply the model tuned on ImageNet to CIFAR10 \cite{krizhevsky2009learning}, CIFAR100 \cite{krizhevsky2009learning}, and four fine-grained datasets \cite{wah2011caltech,bossard2014food,parkhi2012cats,krause20133d}. For CIFAR datasets, we follow the OpenOOD setting \cite{yang2022openood} and use MNIST \cite{deng2012mnist}, SVHN \cite{netzer2011reading}, Places \cite{zhou2017places}, and Textures \cite{cimpoi2014describing} as OOD test datasets.
For fine-grained datasets, we follow \cite{cao2024envisioning} to randomly select half of the classes as ID and use the remaining half as OOD. 
We also evaluate the generalization to unseen image styles, where we apply the model tuned on ImageNet to covariate-shifted ID datasets of ImageNet-R \cite{hendrycks2021many}, ImageNet-V2 \cite{recht2019imagenet}, ImageNet-A \cite{hendrycks2021natural}, and ImageNet-Sketch \cite{wang2019learning}.

\noindent\textbf{Implementation Details.}
We conduct the experiments under the four-shot learning setting ($S$=4) with a ViT-B/16 visual encoder. 
For each training sample, we generate $P$=256 random-sized crops, supplemented with color augmentation methods such as color jitter and grayscale. From these crops, we select $Q$=32 crops to construct the positive training set $\mathcal{D}_p$ and the negative training set $\mathcal{D}_n$, respectively. 
We utilize $10,000$ negative text labels following \cite{jiang2024negative}.
In the optimization process, we use the AdamW optimizer \cite{kingma2014adam} to train the model for three epochs at a learning rate of 1e-5. We use $\lambda_1$=0.3 and $\lambda_2$=100 in all experiments. 
We initialize $\bm{\alpha}$ as all ones, $\bm{\beta}$ as all zeros, and constrain the initial output of $\widehat{h}_{*}(\cdot)$ to be zeros to avoid disturbing pre-trained knowledge at the beginning of training.
\vspace{-0.3cm}

\subsection{Main Results}
\begin{figure*}[t]
    \centering
    \begin{subfigure}[t]{0.25\textwidth}
        \centering
    \includegraphics[width=\textwidth]{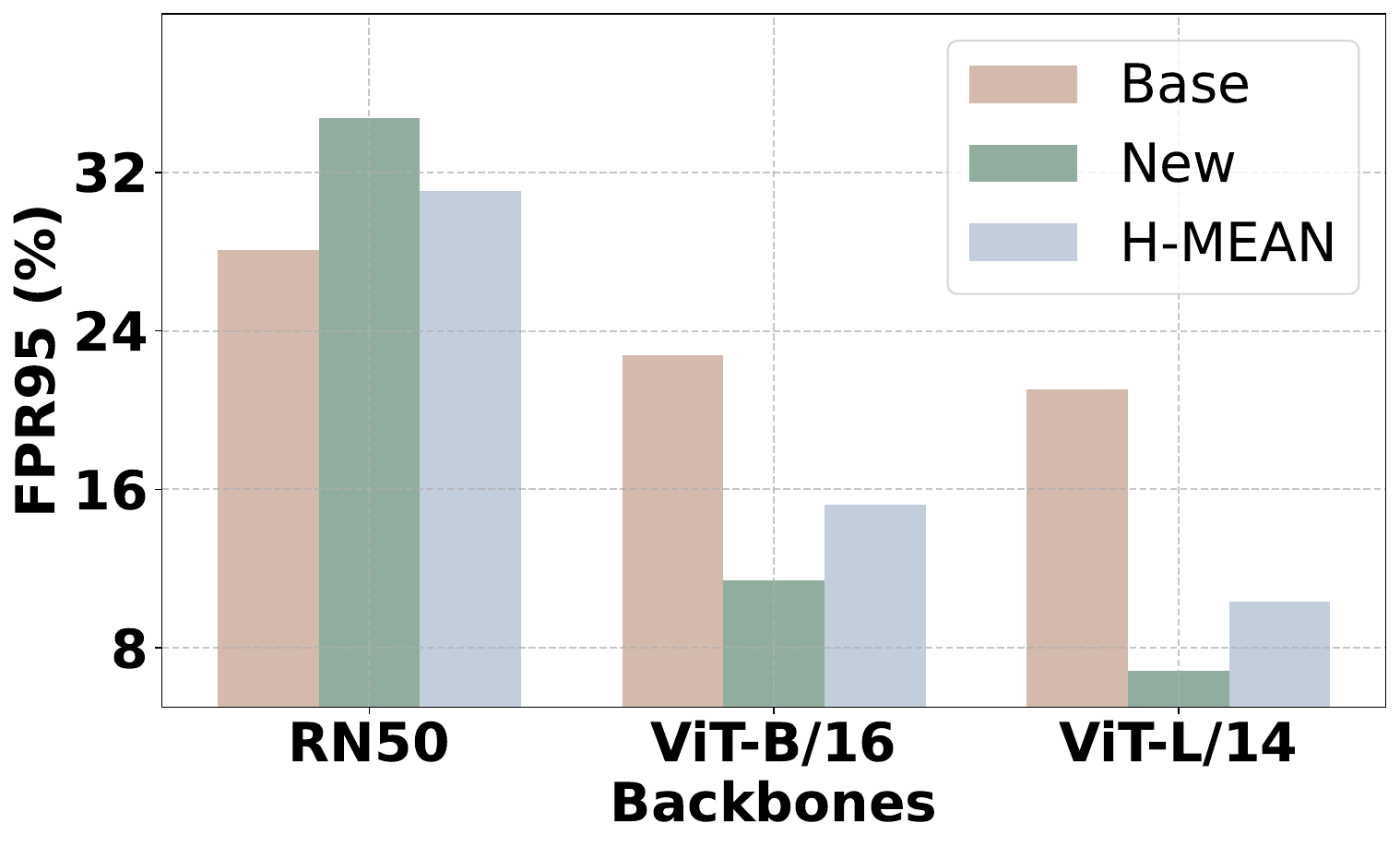}
        \caption{Backbones}
        \label{fig:backbone}
    \end{subfigure}
    \hfill
    \begin{subfigure}[t]{0.25\textwidth}
        \centering
        \includegraphics[width=\textwidth]{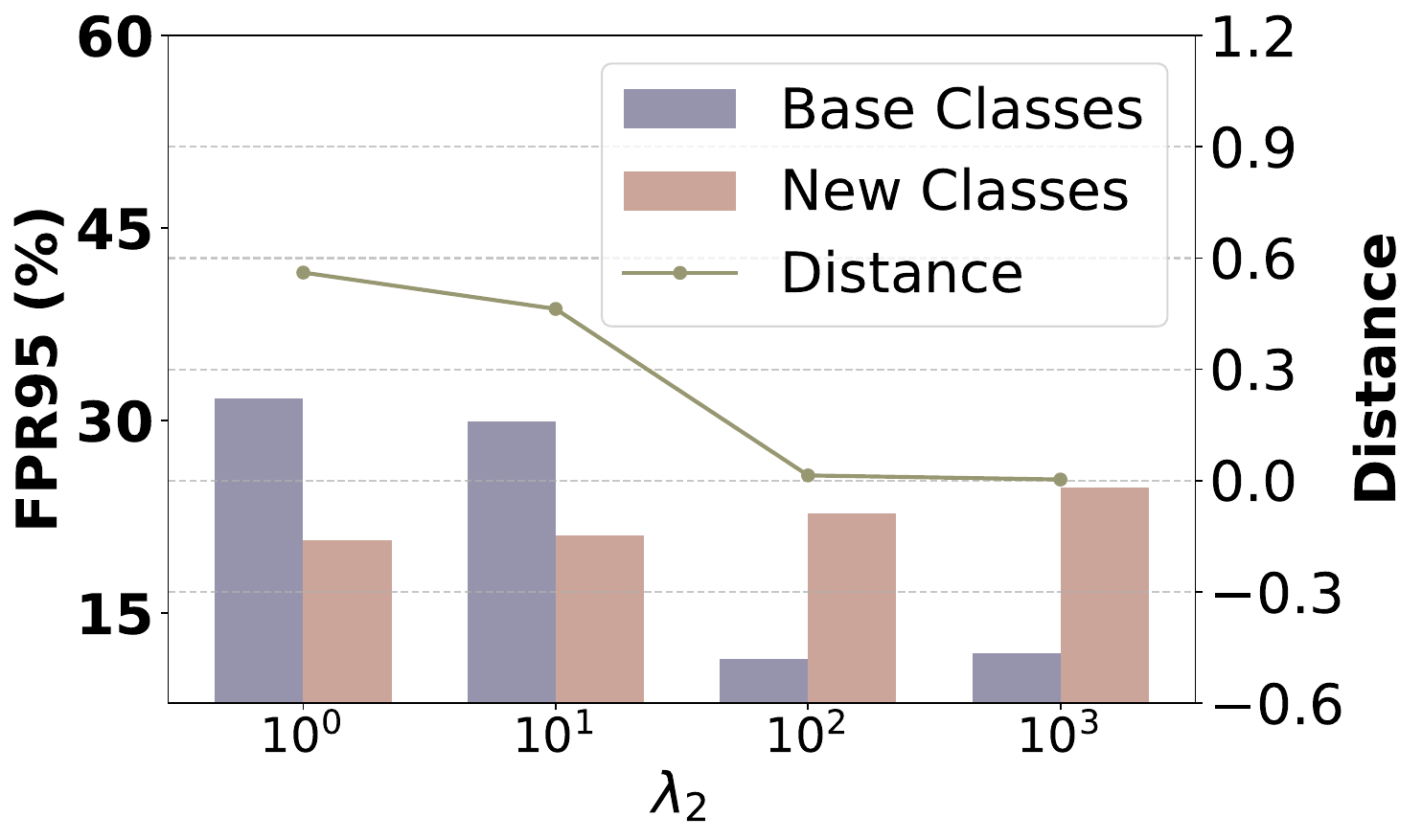}
        \caption{Values of $\lambda_2$}
        \label{fig:distances}
    \end{subfigure}
    \hfill
    \begin{subfigure}[t]{0.24\textwidth}
        \centering
        \includegraphics[width=\textwidth]{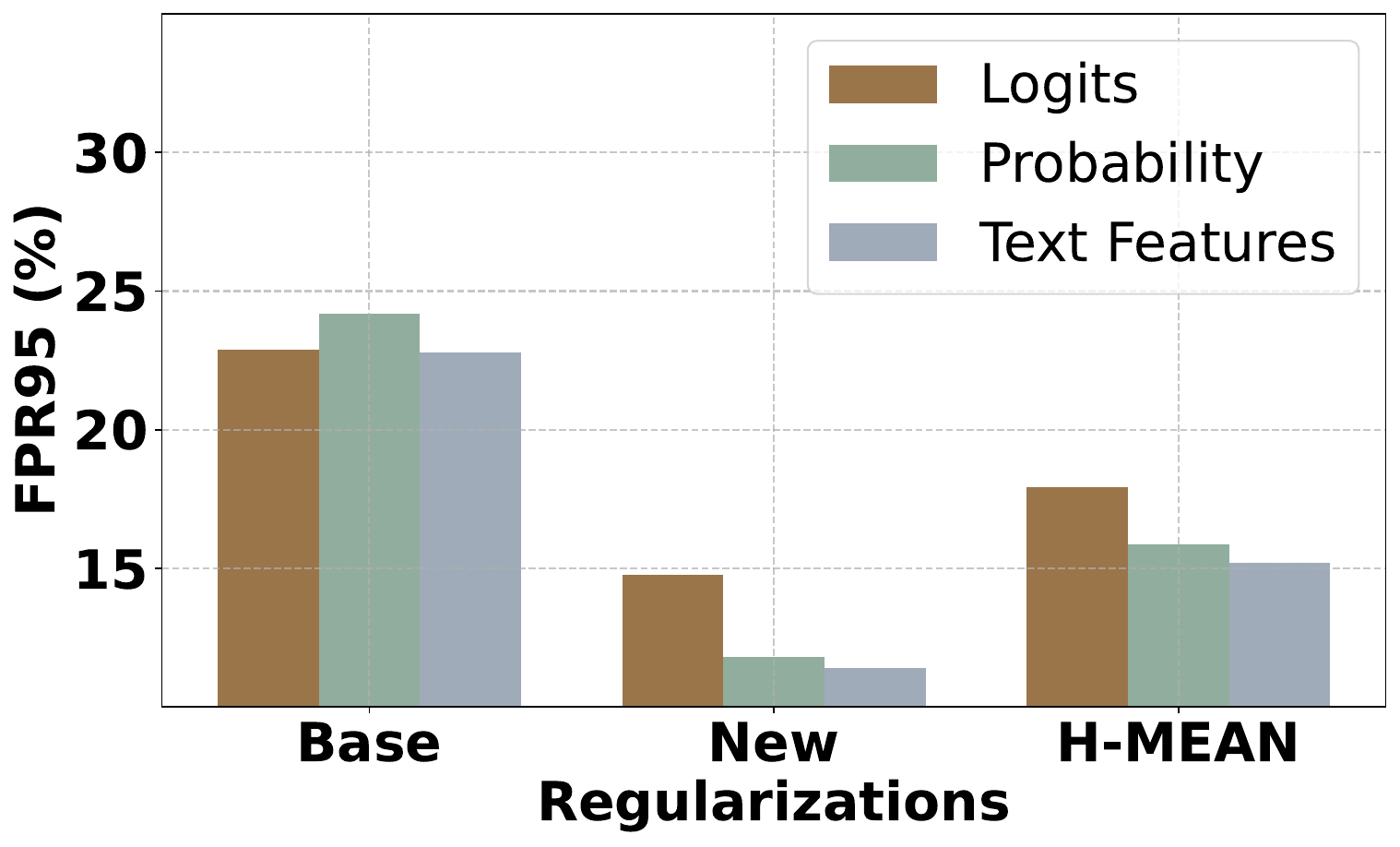}
        \caption{Knowledge Regularization }
        \label{fig:kp}
    \end{subfigure}
    \hfill
    \begin{subfigure}[t]{0.24\textwidth}
        \centering
        \includegraphics[width=\textwidth]{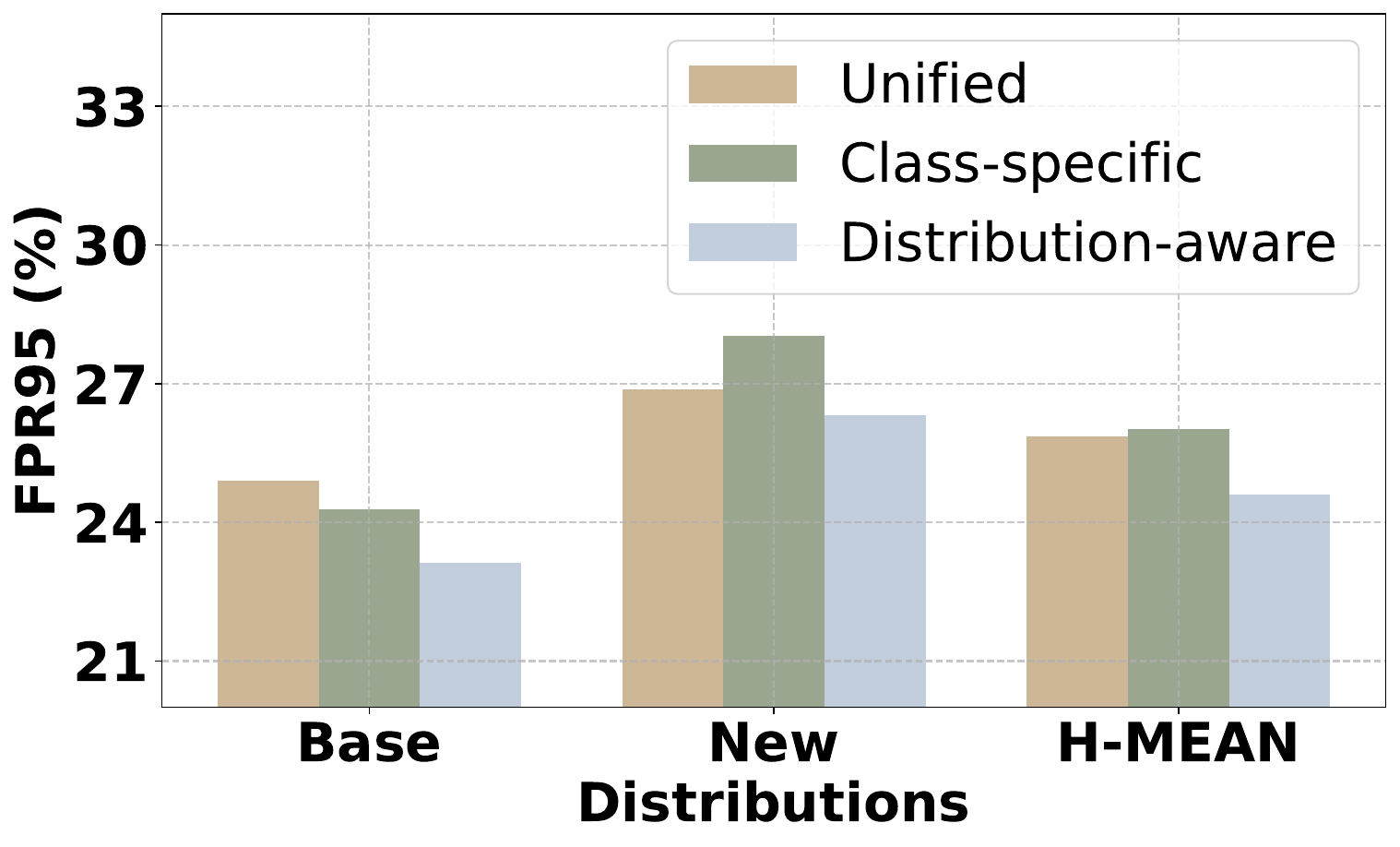}
        \caption{Distributions of NFT}
        \label{fig:distributions}
    \end{subfigure}
    \vspace{-0.3cm}
    \caption{Analyses on (a) hidden dimensions, (b) values of $\lambda_2$, (c) knowledge regularization strategies, and (d) distributions.
    }
    \vspace{-0.3cm}
    \label{fig:Analysis}
\end{figure*}

\noindent\textbf{Results on Training Dataset.}
As shown in the Tab. \ref{tab:four_ood_datasets}, our KR-NFT outperforms other competitors on average. We report the traditional methods \cite{hendrycks2016baseline, liang2017enhancing, liu2020energy, huang2021importance, wang2022vim, sun2022out, du2022unknown, tao2023non} by fine-tuning CLIP visual encoders with the ImageNet training data as described in \cite{jiang2024negative}, and reference the results of other methods \cite{esmaeilpour2022zero, wang2023clipn, cao2024envisioning, nie2024out, miyai2024locoop, bai2024id, li2024learning, zhang2024lapt} directly from their respective papers.
Although setting \(\lambda_2 = 0\) yields superior results on the Base ImageNet dataset, configuring \(\lambda_2\) to a moderate value optimally balances performance across both the Base and New test sets, as analyzed in Fig. \ref{fig:distances}.

\noindent\textbf{Generalization Results to Unseen Classes.}  A good OOD detection model should not only perform well on training data but also demonstrate strong generalization on unseen classes and styles. To this end, we assess the generalization performance of our model by applying it, once tuned on the ImageNet dataset, to the ID datasets of CIFAR10, CIFAR100, and four fine-grained datasets. As shown in the Tab. \ref{tab:cross_datasets}, while existing model tuning methods, such as LoCoOp and LAPT, achieve certain improvements in training classes, their generalization to new classes significantly declines. In contrast, our method not only achieves substantial improvements on training classes but also enhances performance on unseen classes, presenting a comprehensively enhanced OOD detection capability.

\begin{table*}[t] 
\footnotesize
\centering
\begin{tabular}{lcccccccc|cc}
\toprule
\multirow{3}{*}{Methods} & \multicolumn{10}{c}{Unseen New Styles}  \\
 & \multicolumn{2}{c}{ImageNet-S} & \multicolumn{2}{c}{ImageNet-A} & \multicolumn{2}{c}{ImageNet-R} & \multicolumn{2}{c|}{ImageNet-V2} & \multicolumn{2}{c}{Average} \\ 
 \vspace{2pt} 
 & \fontsize{8}{12}\selectfont AUROC$\uparrow$ & \fontsize{8}{12}\selectfont FPR95$\downarrow$& \fontsize{8}{12}\selectfont AUROC$\uparrow$ & \fontsize{8}{12}\selectfont FPR95$\downarrow$& \fontsize{8}{12}\selectfont AUROC$\uparrow$ & \fontsize{8}{12}\selectfont FPR95$\downarrow$& \fontsize{8}{12}\selectfont AUROC$\uparrow$ & \fontsize{8}{12}\selectfont FPR95$\downarrow$ & \fontsize{8}{12}\selectfont AUROC$\uparrow$ & \fontsize{8}{12}\selectfont FPR95$\downarrow$ \\
 \midrule
  \multicolumn{11}{c}{\textbf{Zero-shot methods}} \\
  MCM \cite{ming2022delving} & 82.26 & 70.13 & 72.16 & 80.88 & 76.68 & 76.72 & 87.43 & 55.61 & 79.68 & 70.84 \\
  NegLabel \cite{jiang2024negative} & 93.59 & 27.42 & 87.94 & 43.23 & 94.54 & 20.63 & 93.08 & 29.77 & 92.29 & 30.26 \\
  \midrule
  \multicolumn{11}{c}{\textbf{Tuning-based methods}} \\ 
  LoCoOp \cite{miyai2024locoop} & 68.32 & 73.35 & 72.66 & 77.64 & 50.36 & 89.14 & 74.33 & 62.48 & 66.42 & 76.65 \\
  ID-like \cite{bai2024id} & 74.29 & 75.92 & 80.23 & 67.24 & 83.03 & 61.09 & 88.11 & 49.57 & 81.41 & 63.46 \\
    NegPrompt \cite{li2024learning} & 77.57 & 65.18 & 69.82 & 76.57 & 80.70 & 56.01 & 74.06 & 63.42 & 75.53 & 65.30 \\
  SCT \cite{yu2024self} & 85.11 & 53.30 & 80.30 & 63.13 & 85.23 & 48.81 & 90.66 & 36.47 & 85.33 & 50.43 \\  
  LAPT \cite{zhang2024lapt} & 84.80 & 51.87 & 87.18 & 51.47 & 87.75 & 47.42 & 93.30 & \textbf{28.84} & 88.26 & 44.90 \\
\rowcolor{lightpink}  \textbf{KR-NFT (Ours)} & \textbf{94.10} & \textbf{24.81} & \textbf{89.13} & \textbf{38.46} & \textbf{94.66} & \textbf{20.12} & \textbf{93.61} & 29.13 & \textbf{92.88} & \textbf{28.13} \\
\bottomrule
\end{tabular}
   \caption{OOD detection results on unseen styles, where results of \cite{miyai2024locoop,bai2024id,li2024learning,zhang2024lapt} and ours are reported with models tuned on the ImageNet dataset. Pre-trained CLIP model is utilized in \cite{ming2022delving,jiang2024negative}. Detailed results are provided in the Supplementary Material.
    }
    \vspace{-0.6cm}
    \label{tab:domain_generalization}
\end{table*}

\noindent\textbf{Generalization Results to Unseen Styles.}
In addition to evaluating the generalization performance on unseen classes, we also validate our model on unseen image styles. As shown in Tab. \ref{tab:domain_generalization}, our method demonstrates strong generalization across various natural style shifts compared to other few-shot training methods, confirming its robustness.

\noindent\textbf{Compatibility.}
As shown in Tab. \ref{tab:compatibility}, our method is highly compatible with existing prompt-tuning-based approaches, bringing consistent improvement to OOD detection capabilities.

\noindent\textbf{Evaluation on OpenOOD benchmark.}
We evaluated our method on the OpenOOD benchmark, which includes far-OOD and near-OOD datasets. As shown in the table \ref{tab:openood}, our KR-NFT performs better than Neglabel on both far-OOD and near-OOD datasets. This demonstrates the robustness of KR-NFT, making it applicable across various OOD scenarios.


\vspace{-0.2cm}

\begin{table}[h] 
\small
\begin{tabular}{l|cc|cc}
\toprule
\multirow{2}{*}{Methods} & \multicolumn{2}{c|}{Unseen New Classes} & \multicolumn{2}{c}{Unseen New Styles} \\
 & \fontsize{8}{12}\selectfont AUROC$\uparrow$ & \fontsize{8}{12}\selectfont FPR95$\downarrow$ & \fontsize{8}{12}\selectfont AUROC$\uparrow$ & \fontsize{8}{12}\selectfont FPR95$\downarrow$ \\
\midrule
LoCoOp \cite{miyai2024locoop} & 82.82 & 52.21 & 66.42 & 76.65 \\
LoCoOp + KR-NFT \cite{miyai2024locoop} & \textbf{85.30} & \textbf{49.37} & \textbf{78.98} & \textbf{70.06} \\
\midrule
SCT \cite{yu2024self} & 85.28 & 48.85 & 85.33 & 50.43 \\ 
SCT + KR-NFT \cite{yu2024self} & \textbf{86.01} & \textbf{45.53} & \textbf{85.96} & \textbf{49.58} \\ 
\midrule
LAPT \cite{zhang2024lapt} & 87.39 & 40.43 & 88.26 & 44.90 \\
LAPT + KR-NFT \cite{li2024learning} & \textbf{88.57} & \textbf{36.35} & \textbf{89.68} & \textbf{41.08} \\ 
\bottomrule
\end{tabular}
\caption{OOD detection performance on compatibility experiments. We report the average OOD detection results on unseen classes and style.}
\vspace{-0.5cm}
\label{tab:compatibility}
\end{table}

\vspace{-0.4cm}

\begin{table}[h]
\centering
\small    
\vspace{-0.2cm}
\begin{tabular}{l|c@{\hskip 2pt}c|c@{\hskip 2pt}c|}
\toprule
\multirow{2}{*}{ Methods } & \multicolumn{2}{|c|}{ FPR95 $\downarrow$} & \multicolumn{2}{|c|}{$\mathrm{AUROC} \uparrow$} \\
\cline{2-5}
& NearOOD & FarOOD & NearOOD & FarOOD  \\
\midrule
GEN \cite{liu2023gen}  & -- & -- & 78.97 & 90.98 \\
AugMix \cite{hendrycks2019augmix} + ReAct \cite{sun2021react} & -- & -- & 79.94 & 93.70  \\
RMDS \cite{ren2021simple} & -- & -- & 80.09 & 92.60  \\
AugMix \cite{hendrycks2019augmix} + ASH \cite{djurisic2022extremely}   & -- & -- & 82.16 & 96.05 \\
\midrule
MCM \cite{ming2022delving}        & 79.02 & 68.54 & 60.11 & 84.77 \\
NegLabel \cite{jiang2024negative} & 68.18 & 25.40 & 76.92 & 93.30 \\
\rowcolor{lightpink} 
Ours & 67.24 & 19.08 & 77.73 & 95.82 \\
\bottomrule
\end{tabular}
\caption{OOD detection results on the OpenOOD benchmark, where ImageNet-1k is adopted as ID dataset. Full results are available in the Supplementary Materials.}\label{tab:openood}
\vspace{-0.8cm}
\end{table}


\subsection{Ablation and Analyses}

\subsubsection{Ablation Study.}   
Our analyses are mainly conducted under the cross-class generalization setting, where we tune a pre-trained VLM on ImageNet and validate its performance on both ImageNet (Base) and the unseen CIFAR10 (New).
We introduce H-MEAN, a harmonic mean of results from both settings, to assess the comprehensive OOD detection capability.

\noindent\textbf{Component Ablation.}
As illustrated in Tab. \ref{tab:ablation}, adapting text features with the transformation function indeed improved performance on the Base classes, but significantly compromised generalization to unseen New classes. Introducing the image-conditional transformation mitigated the performance decline in New classes, but there still remains a gap compared to the pre-trained CLIP, suggesting a forgetting of pre-trained knowledge. The inclusion of an knowledge regularization objective $\mathcal{L}_{kr}$ successfully balanced learning for the new task with the preservation of pre-trained knowledge, achieving results that surpass the pre-trained CLIP in both Base and New settings. Lastly, incorporating distribution-aware information further enhanced model performance, leading to the best results.

\vspace{-0.2cm}
\begin{table}[ht]
\footnotesize
    \centering
        \begin{tabular}{@{}cccc|c|c|c@{}}
        \toprule
        \multicolumn{4}{c|}{Components} & \multicolumn{3}{c}{FRR95 $\downarrow$}  \\
        FT  & CoFT & KR & NFT & Base & New & H-MEAN  \\
        \midrule
        \multicolumn{4}{c|}{NegLabel with Pre-trained CLIP} & 25.40 & 12.15 & 16.45 \\
        \midrule
        \Checkmark & \textcolor{gray}{\XSolidBrush} & \textcolor{gray}{\XSolidBrush} & \textcolor{gray}{\XSolidBrush} & 25.02 & 26.08 & 25.54  \\
        \Checkmark & \Checkmark & \textcolor{gray}{\XSolidBrush} & \textcolor{gray}{\XSolidBrush} & 25.21 & 20.43 & 22.57 \\
        \Checkmark & \Checkmark & \Checkmark & \textcolor{gray}{\XSolidBrush} & 25.09 & 11.86 &
        16.11 \\
        \Checkmark & \Checkmark & \Checkmark & \Checkmark & 22.79 & 11.41 & 15.20 \\
        \bottomrule
        \end{tabular}
        \caption{Ablation results on the proposed components, where `FT' indicates the feature tuning in Eq. (\ref{equ:scale_shift}), `CoFT' denotes the image-conditional feature tuning variant in Eq. (\ref{equ:conditional_scale_shift}), `KR' is the pre-trained feature knowledge regularization loss in Eq. (\ref{eq:consistency_regularization}), and `NFT' indicates the negative feature tuning in Eq. (\ref{eq:distribution_aware_transformation}). 
        }
        \vspace{-0.4cm}
        \label{tab:ablation}
\end{table}

\noindent\textbf{Different Backbones.}
Our KR-NFT is compatible with different VLM backbones. As shown in the Fig. \ref{fig:backbone}, KR-NFT achieves a lower FPR95 on larger backbones across various OOD detection scenarios, indicating that larger backbones provide greater performance improvement. Furthermore, compared to Neglabel, KR-NFT demonstrates better performance across different backbones, highlighting the strong robustness of KR-NFT with various architectures.

\noindent\textbf{The sensitivity to different $\lambda_2$.} 
As shown in Fig. \ref{fig:distances}, with the increase of $\lambda_2$, the performance on the Base dataset tends to decrease, as this regularization limits learning on the Base classes. However, a higher value of \(\lambda_2\) is beneficial for the preservation of pre-trained knowledge, by reducing the distance between the pre-trained text features and new learned ones, as validated by the improved performance on the New classes dataset. We find that a moderate value of \(\lambda_2=100\) achieves good results on both datasets, balancing the preservation of pre-trained knowledge and acquisition of new information. 

\noindent\textbf{Knowledge Regularization Implementations.}
Besides applying regularization on text features $\mC'$ as in Eq. \ref{eq:consistency_regularization}, we also explore the knowledge regularization objective on the logits (\eg, $\mC'\vv$) and probabilities (\eg, Softmax($\mC'\vv$)), which are detailed in the Supplementary Material. As shown in Fig. \ref{fig:kp}, feature regularization yields the best results.

\noindent\textbf{Distribution-aware Transformation.}
As shown in Fig. \ref{fig:distributions}, using distribution-aware transformation as in Eq. \ref{eq:distribution_aware_transformation} outperforms using one unified and class-specific transformation, which is expected as binary transformations can better capture the characteristics of the distributions of ID and OOD images.

\subsubsection{The Effectiveness of KR-NFT}
To understand the effectiveness of KR-NFT, we calculated the cosine similarity between the ID/OOD images and the positive/negative text features with features of pre-trained CLIP and our KR-NFT, the detailed results can be found in the supplementary materials.

\subsubsection{Computation Efficiency.}
As shown in Tab. \ref{tab:efficiency}, although we introduce certain training parameters, our KR-NFT enjoys fast training and small TFLOPs, because we do not need gradient backpropagation through a heavy text encoder like other prompt tuning-based methods. We also observed that although ConNegPT improves generalization on the New dataset compared to NegPT, it significantly slows down the test speed, because it requires re-forwarding the text encoder for each test image. In contrast, our method maintains a fast test speed and outperforms in performance, presenting advantages in both effectiveness and efficiency.

\vspace{-0.2cm}
\begin{table}[ht]
\footnotesize
    \centering
    \begin{tabular}{l|cccccc}
    \toprule
        Methods & Train & Test & TFLOPs & Param. & Base & New \\
        \midrule 
         LoCoOp \cite{miyai2024locoop} & 0.33h & 10.9ms & 38.25 & 8K & 42.32 & 17.35 \\
         ID-like \cite{bai2024id} & 3.3h & 7.4ms & 39.12 & 16K & 31.78 & 54.99 \\
         \midrule
         NegPT$^*$ & 1.60h & 21.5ms & 193.76 & 2K & 24.21 & 27.33 \\
         ConNegPT$^*$ & 1.65h & 936.1ms & 194.53 & 3K & 27.27 & 25.42 \\
         \textbf{KR-NFT (Ours)} & 0.25h & 21.6ms & 0.11 & 6K & 22.79 & 11.41 \\
    \bottomrule
    \end{tabular}
    \vspace{0.2cm}
    \caption{`Train’ and `Test' measures the training time and testing time, respectively. `TFLOPs’ are calculated during training with gradient back-propagation, and `Param.’ presents the number of learnable parameters. `Base' and `New' indicate the FPR95 on the ImageNet and CIFAR10 datasets, respectively. 
    $^*$We compare our KR-NFT with negative prompt tuning (NegPT) and conditional negative prompt tuning (ConNegPT). Results are reported with a GeForce RTX 3090 GPU.
    }
     \vspace{-0.8cm}
    \label{tab:efficiency}
\end{table}
\label{sec:experiments}

\section{Conclusion}
In this paper, we reveal a limitation in existing negative prompt tuning methods for OOD detection: these models suffer from reduced generalization performance on unseen classes and styles. To address this issue, we propose a novel framework named Knowledge Regularized Negative Feature Tuning (KR-NFT). NFT optimizes positive and negative features into distinct spaces, maximizing the separation between ID and OOD images. The image-conditional learnable factors explore instance-adaptive knowledge, reducing overfitting to the training class and style. KR minimizes the discrepancy between pre-trained text features and tuned ones, alleviating knowledge forgetting. Extensive experiments demonstrate the effectiveness of our KR-NFT and its compatibility with other negative prompt-tuning methods. 

\clearpage


\bibliographystyle{ACM-Reference-Format}
\bibliography{sample-base}


\begin{thebibliography}{86}


\ifx \showCODEN    \undefined \def \showCODEN     #1{\unskip}     \fi
\ifx \showISBNx    \undefined \def \showISBNx     #1{\unskip}     \fi
\ifx \showISBNxiii \undefined \def \showISBNxiii  #1{\unskip}     \fi
\ifx \showISSN     \undefined \def \showISSN      #1{\unskip}     \fi
\ifx \showLCCN     \undefined \def \showLCCN      #1{\unskip}     \fi
\ifx \shownote     \undefined \def \shownote      #1{#1}          \fi
\ifx \showarticletitle \undefined \def \showarticletitle #1{#1}   \fi
\ifx \showURL      \undefined \def \showURL       {\relax}        \fi
\providecommand\bibfield[2]{#2}
\providecommand\bibinfo[2]{#2}
\providecommand\natexlab[1]{#1}
\providecommand\showeprint[2][]{arXiv:#2}

\bibitem[Abati et~al\mbox{.}(2019)]%
        {abati2019latent}
\bibfield{author}{\bibinfo{person}{Davide Abati}, \bibinfo{person}{Angelo Porrello}, \bibinfo{person}{Simone Calderara}, {and} \bibinfo{person}{Rita Cucchiara}.} \bibinfo{year}{2019}\natexlab{}.
\newblock \showarticletitle{Latent space autoregression for novelty detection}. In \bibinfo{booktitle}{\emph{Proceedings of the IEEE/CVF conference on computer vision and pattern recognition}}. \bibinfo{pages}{481--490}.
\newblock


\bibitem[Anwar et~al\mbox{.}(2018)]%
        {anwar2018medical}
\bibfield{author}{\bibinfo{person}{Syed~Muhammad Anwar}, \bibinfo{person}{Muhammad Majid}, \bibinfo{person}{Adnan Qayyum}, \bibinfo{person}{Muhammad Awais}, \bibinfo{person}{Majdi Alnowami}, {and} \bibinfo{person}{Muhammad~Khurram Khan}.} \bibinfo{year}{2018}\natexlab{}.
\newblock \showarticletitle{Medical image analysis using convolutional neural networks: a review}.
\newblock \bibinfo{journal}{\emph{Journal of medical systems}}  \bibinfo{volume}{42} (\bibinfo{year}{2018}), \bibinfo{pages}{1--13}.
\newblock


\bibitem[Bai et~al\mbox{.}(2024)]%
        {bai2024id}
\bibfield{author}{\bibinfo{person}{Yichen Bai}, \bibinfo{person}{Zongbo Han}, \bibinfo{person}{Bing Cao}, \bibinfo{person}{Xiaoheng Jiang}, \bibinfo{person}{Qinghua Hu}, {and} \bibinfo{person}{Changqing Zhang}.} \bibinfo{year}{2024}\natexlab{}.
\newblock \showarticletitle{ID-like Prompt Learning for Few-Shot Out-of-Distribution Detection}. In \bibinfo{booktitle}{\emph{Proceedings of the IEEE/CVF Conference on Computer Vision and Pattern Recognition}}. \bibinfo{pages}{17480--17489}.
\newblock


\bibitem[Bogdoll et~al\mbox{.}(2022)]%
        {bogdoll2022anomaly}
\bibfield{author}{\bibinfo{person}{Daniel Bogdoll}, \bibinfo{person}{Maximilian Nitsche}, {and} \bibinfo{person}{J~Marius Z{\"o}llner}.} \bibinfo{year}{2022}\natexlab{}.
\newblock \showarticletitle{Anomaly detection in autonomous driving: A survey}. In \bibinfo{booktitle}{\emph{Proceedings of the IEEE/CVF conference on computer vision and pattern recognition}}. \bibinfo{pages}{4488--4499}.
\newblock


\bibitem[Bossard et~al\mbox{.}(2014)]%
        {bossard2014food}
\bibfield{author}{\bibinfo{person}{Lukas Bossard}, \bibinfo{person}{Matthieu Guillaumin}, {and} \bibinfo{person}{Luc Van~Gool}.} \bibinfo{year}{2014}\natexlab{}.
\newblock \showarticletitle{Food-101--mining discriminative components with random forests}. In \bibinfo{booktitle}{\emph{Computer vision--ECCV 2014: 13th European conference, zurich, Switzerland, September 6-12, 2014, proceedings, part VI 13}}. Springer, \bibinfo{pages}{446--461}.
\newblock


\bibitem[Cao et~al\mbox{.}(2024)]%
        {cao2024envisioning}
\bibfield{author}{\bibinfo{person}{Chentao Cao}, \bibinfo{person}{Zhun Zhong}, \bibinfo{person}{Zhanke Zhou}, \bibinfo{person}{Yang Liu}, \bibinfo{person}{Tongliang Liu}, {and} \bibinfo{person}{Bo Han}.} \bibinfo{year}{2024}\natexlab{}.
\newblock \showarticletitle{Envisioning Outlier Exposure by Large Language Models for Out-of-Distribution Detection}.
\newblock \bibinfo{journal}{\emph{arXiv preprint arXiv:2406.00806}} (\bibinfo{year}{2024}).
\newblock


\bibitem[Cimpoi et~al\mbox{.}(2014)]%
        {cimpoi2014describing}
\bibfield{author}{\bibinfo{person}{Mircea Cimpoi}, \bibinfo{person}{Subhransu Maji}, \bibinfo{person}{Iasonas Kokkinos}, \bibinfo{person}{Sammy Mohamed}, {and} \bibinfo{person}{Andrea Vedaldi}.} \bibinfo{year}{2014}\natexlab{}.
\newblock \showarticletitle{Describing textures in the wild}. In \bibinfo{booktitle}{\emph{Proceedings of the IEEE conference on computer vision and pattern recognition}}. \bibinfo{pages}{3606--3613}.
\newblock


\bibitem[Deng et~al\mbox{.}(2009)]%
        {deng2009imagenet}
\bibfield{author}{\bibinfo{person}{Jia Deng}, \bibinfo{person}{Wei Dong}, \bibinfo{person}{Richard Socher}, \bibinfo{person}{Li-Jia Li}, \bibinfo{person}{Kai Li}, {and} \bibinfo{person}{Li Fei-Fei}.} \bibinfo{year}{2009}\natexlab{}.
\newblock \showarticletitle{Imagenet: A large-scale hierarchical image database}. In \bibinfo{booktitle}{\emph{2009 IEEE conference on computer vision and pattern recognition}}. Ieee, \bibinfo{pages}{248--255}.
\newblock


\bibitem[Deng(2012)]%
        {deng2012mnist}
\bibfield{author}{\bibinfo{person}{Li Deng}.} \bibinfo{year}{2012}\natexlab{}.
\newblock \showarticletitle{The mnist database of handwritten digit images for machine learning research [best of the web]}.
\newblock \bibinfo{journal}{\emph{IEEE signal processing magazine}} \bibinfo{volume}{29}, \bibinfo{number}{6} (\bibinfo{year}{2012}), \bibinfo{pages}{141--142}.
\newblock


\bibitem[Derakhshani et~al\mbox{.}(2023)]%
        {derakhshani2023bayesian}
\bibfield{author}{\bibinfo{person}{Mohammad~Mahdi Derakhshani}, \bibinfo{person}{Enrique Sanchez}, \bibinfo{person}{Adrian Bulat}, \bibinfo{person}{Victor G~Turrisi da Costa}, \bibinfo{person}{Cees~GM Snoek}, \bibinfo{person}{Georgios Tzimiropoulos}, {and} \bibinfo{person}{Brais Martinez}.} \bibinfo{year}{2023}\natexlab{}.
\newblock \showarticletitle{Bayesian prompt learning for image-language model generalization}. In \bibinfo{booktitle}{\emph{Proceedings of the IEEE/CVF International Conference on Computer Vision}}. \bibinfo{pages}{15237--15246}.
\newblock


\bibitem[DeVries and Taylor(2018)]%
        {devries2018learning}
\bibfield{author}{\bibinfo{person}{Terrance DeVries} {and} \bibinfo{person}{Graham~W Taylor}.} \bibinfo{year}{2018}\natexlab{}.
\newblock \showarticletitle{Learning confidence for out-of-distribution detection in neural networks}.
\newblock \bibinfo{journal}{\emph{arXiv preprint arXiv:1802.04865}} (\bibinfo{year}{2018}).
\newblock


\bibitem[Djurisic et~al\mbox{.}(2022)]%
        {djurisic2022extremely}
\bibfield{author}{\bibinfo{person}{Andrija Djurisic}, \bibinfo{person}{Nebojsa Bozanic}, \bibinfo{person}{Arjun Ashok}, {and} \bibinfo{person}{Rosanne Liu}.} \bibinfo{year}{2022}\natexlab{}.
\newblock \showarticletitle{Extremely simple activation shaping for out-of-distribution detection}.
\newblock \bibinfo{journal}{\emph{arXiv preprint arXiv:2209.09858}} (\bibinfo{year}{2022}).
\newblock


\bibitem[Dong et~al\mbox{.}(2022)]%
        {dong2022neural}
\bibfield{author}{\bibinfo{person}{Xin Dong}, \bibinfo{person}{Junfeng Guo}, \bibinfo{person}{Ang Li}, \bibinfo{person}{Wei-Te Ting}, \bibinfo{person}{Cong Liu}, {and} \bibinfo{person}{HT Kung}.} \bibinfo{year}{2022}\natexlab{}.
\newblock \showarticletitle{Neural mean discrepancy for efficient out-of-distribution detection}. In \bibinfo{booktitle}{\emph{Proceedings of the IEEE/CVF Conference on Computer Vision and Pattern Recognition}}. \bibinfo{pages}{19217--19227}.
\newblock


\bibitem[Du et~al\mbox{.}(2022)]%
        {du2022unknown}
\bibfield{author}{\bibinfo{person}{Xuefeng Du}, \bibinfo{person}{Xin Wang}, \bibinfo{person}{Gabriel Gozum}, {and} \bibinfo{person}{Yixuan Li}.} \bibinfo{year}{2022}\natexlab{}.
\newblock \showarticletitle{Unknown-aware object detection: Learning what you don't know from videos in the wild}. In \bibinfo{booktitle}{\emph{Proceedings of the IEEE/CVF Conference on Computer Vision and Pattern Recognition}}. \bibinfo{pages}{13678--13688}.
\newblock


\bibitem[Esmaeilpour et~al\mbox{.}(2022)]%
        {esmaeilpour2022zero}
\bibfield{author}{\bibinfo{person}{Sepideh Esmaeilpour}, \bibinfo{person}{Bing Liu}, \bibinfo{person}{Eric Robertson}, {and} \bibinfo{person}{Lei Shu}.} \bibinfo{year}{2022}\natexlab{}.
\newblock \showarticletitle{Zero-shot out-of-distribution detection based on the pre-trained model clip}. In \bibinfo{booktitle}{\emph{Proceedings of the AAAI conference on artificial intelligence}}, Vol.~\bibinfo{volume}{36}. \bibinfo{pages}{6568--6576}.
\newblock


\bibitem[Fort et~al\mbox{.}(2021)]%
        {fort2021exploring}
\bibfield{author}{\bibinfo{person}{Stanislav Fort}, \bibinfo{person}{Jie Ren}, {and} \bibinfo{person}{Balaji Lakshminarayanan}.} \bibinfo{year}{2021}\natexlab{}.
\newblock \showarticletitle{Exploring the limits of out-of-distribution detection}.
\newblock \bibinfo{journal}{\emph{Advances in Neural Information Processing Systems}}  \bibinfo{volume}{34} (\bibinfo{year}{2021}), \bibinfo{pages}{7068--7081}.
\newblock


\bibitem[Gao et~al\mbox{.}(2024)]%
        {gao2024clip}
\bibfield{author}{\bibinfo{person}{Peng Gao}, \bibinfo{person}{Shijie Geng}, \bibinfo{person}{Renrui Zhang}, \bibinfo{person}{Teli Ma}, \bibinfo{person}{Rongyao Fang}, \bibinfo{person}{Yongfeng Zhang}, \bibinfo{person}{Hongsheng Li}, {and} \bibinfo{person}{Yu Qiao}.} \bibinfo{year}{2024}\natexlab{}.
\newblock \showarticletitle{Clip-adapter: Better vision-language models with feature adapters}.
\newblock \bibinfo{journal}{\emph{International Journal of Computer Vision}} \bibinfo{volume}{132}, \bibinfo{number}{2} (\bibinfo{year}{2024}), \bibinfo{pages}{581--595}.
\newblock


\bibitem[He et~al\mbox{.}(2016)]%
        {he2016deep}
\bibfield{author}{\bibinfo{person}{Kaiming He}, \bibinfo{person}{Xiangyu Zhang}, \bibinfo{person}{Shaoqing Ren}, {and} \bibinfo{person}{Jian Sun}.} \bibinfo{year}{2016}\natexlab{}.
\newblock \showarticletitle{Deep residual learning for image recognition}. In \bibinfo{booktitle}{\emph{Proceedings of the IEEE conference on computer vision and pattern recognition}}. \bibinfo{pages}{770--778}.
\newblock


\bibitem[Hendrycks et~al\mbox{.}(2021a)]%
        {hendrycks2021many}
\bibfield{author}{\bibinfo{person}{Dan Hendrycks}, \bibinfo{person}{Steven Basart}, \bibinfo{person}{Norman Mu}, \bibinfo{person}{Saurav Kadavath}, \bibinfo{person}{Frank Wang}, \bibinfo{person}{Evan Dorundo}, \bibinfo{person}{Rahul Desai}, \bibinfo{person}{Tyler Zhu}, \bibinfo{person}{Samyak Parajuli}, \bibinfo{person}{Mike Guo}, {et~al\mbox{.}}} \bibinfo{year}{2021}\natexlab{a}.
\newblock \showarticletitle{The many faces of robustness: A critical analysis of out-of-distribution generalization}. In \bibinfo{booktitle}{\emph{Proceedings of the IEEE/CVF international conference on computer vision}}. \bibinfo{pages}{8340--8349}.
\newblock


\bibitem[Hendrycks and Gimpel(2016)]%
        {hendrycks2016baseline}
\bibfield{author}{\bibinfo{person}{Dan Hendrycks} {and} \bibinfo{person}{Kevin Gimpel}.} \bibinfo{year}{2016}\natexlab{}.
\newblock \showarticletitle{A baseline for detecting misclassified and out-of-distribution examples in neural networks}.
\newblock \bibinfo{journal}{\emph{arXiv preprint arXiv:1610.02136}} (\bibinfo{year}{2016}).
\newblock


\bibitem[Hendrycks et~al\mbox{.}(2020)]%
        {hendrycks2020pretrained}
\bibfield{author}{\bibinfo{person}{Dan Hendrycks}, \bibinfo{person}{Xiaoyuan Liu}, \bibinfo{person}{Eric Wallace}, \bibinfo{person}{Adam Dziedzic}, \bibinfo{person}{Rishabh Krishnan}, {and} \bibinfo{person}{Dawn Song}.} \bibinfo{year}{2020}\natexlab{}.
\newblock \showarticletitle{Pretrained transformers improve out-of-distribution robustness}.
\newblock \bibinfo{journal}{\emph{arXiv preprint arXiv:2004.06100}} (\bibinfo{year}{2020}).
\newblock


\bibitem[Hendrycks et~al\mbox{.}(2018)]%
        {hendrycks2018deep}
\bibfield{author}{\bibinfo{person}{Dan Hendrycks}, \bibinfo{person}{Mantas Mazeika}, {and} \bibinfo{person}{Thomas Dietterich}.} \bibinfo{year}{2018}\natexlab{}.
\newblock \showarticletitle{Deep anomaly detection with outlier exposure}.
\newblock \bibinfo{journal}{\emph{arXiv preprint arXiv:1812.04606}} (\bibinfo{year}{2018}).
\newblock


\bibitem[Hendrycks et~al\mbox{.}(2019)]%
        {hendrycks2019augmix}
\bibfield{author}{\bibinfo{person}{Dan Hendrycks}, \bibinfo{person}{Norman Mu}, \bibinfo{person}{Ekin~D Cubuk}, \bibinfo{person}{Barret Zoph}, \bibinfo{person}{Justin Gilmer}, {and} \bibinfo{person}{Balaji Lakshminarayanan}.} \bibinfo{year}{2019}\natexlab{}.
\newblock \showarticletitle{Augmix: A simple data processing method to improve robustness and uncertainty}.
\newblock \bibinfo{journal}{\emph{arXiv preprint arXiv:1912.02781}} (\bibinfo{year}{2019}).
\newblock


\bibitem[Hendrycks et~al\mbox{.}(2021b)]%
        {hendrycks2021natural}
\bibfield{author}{\bibinfo{person}{Dan Hendrycks}, \bibinfo{person}{Kevin Zhao}, \bibinfo{person}{Steven Basart}, \bibinfo{person}{Jacob Steinhardt}, {and} \bibinfo{person}{Dawn Song}.} \bibinfo{year}{2021}\natexlab{b}.
\newblock \showarticletitle{Natural adversarial examples}. In \bibinfo{booktitle}{\emph{Proceedings of the IEEE/CVF conference on computer vision and pattern recognition}}. \bibinfo{pages}{15262--15271}.
\newblock


\bibitem[Hinton(2015)]%
        {hinton2015distilling}
\bibfield{author}{\bibinfo{person}{Geoffrey Hinton}.} \bibinfo{year}{2015}\natexlab{}.
\newblock \showarticletitle{Distilling the Knowledge in a Neural Network}.
\newblock \bibinfo{journal}{\emph{arXiv preprint arXiv:1503.02531}} (\bibinfo{year}{2015}).
\newblock


\bibitem[Huang et~al\mbox{.}(2021)]%
        {huang2021importance}
\bibfield{author}{\bibinfo{person}{Rui Huang}, \bibinfo{person}{Andrew Geng}, {and} \bibinfo{person}{Yixuan Li}.} \bibinfo{year}{2021}\natexlab{}.
\newblock \showarticletitle{On the importance of gradients for detecting distributional shifts in the wild}.
\newblock \bibinfo{journal}{\emph{Advances in Neural Information Processing Systems}}  \bibinfo{volume}{34} (\bibinfo{year}{2021}), \bibinfo{pages}{677--689}.
\newblock


\bibitem[Huang and Li(2021)]%
        {huang2021mos}
\bibfield{author}{\bibinfo{person}{Rui Huang} {and} \bibinfo{person}{Yixuan Li}.} \bibinfo{year}{2021}\natexlab{}.
\newblock \showarticletitle{Mos: Towards scaling out-of-distribution detection for large semantic space}. In \bibinfo{booktitle}{\emph{Proceedings of the IEEE/CVF Conference on Computer Vision and Pattern Recognition}}. \bibinfo{pages}{8710--8719}.
\newblock


\bibitem[Jiang et~al\mbox{.}(2021)]%
        {jiang2021revisiting}
\bibfield{author}{\bibinfo{person}{Dihong Jiang}, \bibinfo{person}{Sun Sun}, {and} \bibinfo{person}{Yaoliang Yu}.} \bibinfo{year}{2021}\natexlab{}.
\newblock \showarticletitle{Revisiting flow generative models for out-of-distribution detection}. In \bibinfo{booktitle}{\emph{International Conference on Learning Representations}}.
\newblock


\bibitem[Jiang et~al\mbox{.}(2023)]%
        {jiang2023detecting}
\bibfield{author}{\bibinfo{person}{Xue Jiang}, \bibinfo{person}{Feng Liu}, \bibinfo{person}{Zhen Fang}, \bibinfo{person}{Hong Chen}, \bibinfo{person}{Tongliang Liu}, \bibinfo{person}{Feng Zheng}, {and} \bibinfo{person}{Bo Han}.} \bibinfo{year}{2023}\natexlab{}.
\newblock \showarticletitle{Detecting out-of-distribution data through in-distribution class prior}. In \bibinfo{booktitle}{\emph{International Conference on Machine Learning}}. PMLR, \bibinfo{pages}{15067--15088}.
\newblock


\bibitem[Jiang et~al\mbox{.}(2024)]%
        {jiang2024negative}
\bibfield{author}{\bibinfo{person}{Xue Jiang}, \bibinfo{person}{Feng Liu}, \bibinfo{person}{Zhen Fang}, \bibinfo{person}{Hong Chen}, \bibinfo{person}{Tongliang Liu}, \bibinfo{person}{Feng Zheng}, {and} \bibinfo{person}{Bo Han}.} \bibinfo{year}{2024}\natexlab{}.
\newblock \showarticletitle{Negative label guided ood detection with pretrained vision-language models}.
\newblock \bibinfo{journal}{\emph{arXiv preprint arXiv:2403.20078}} (\bibinfo{year}{2024}).
\newblock


\bibitem[Jin et~al\mbox{.}(2022)]%
        {jin2022towards}
\bibfield{author}{\bibinfo{person}{Di Jin}, \bibinfo{person}{Shuyang Gao}, \bibinfo{person}{Seokhwan Kim}, \bibinfo{person}{Yang Liu}, {and} \bibinfo{person}{Dilek Hakkani-T{\"u}r}.} \bibinfo{year}{2022}\natexlab{}.
\newblock \showarticletitle{Towards textual out-of-domain detection without in-domain labels}.
\newblock \bibinfo{journal}{\emph{IEEE/ACM Transactions on Audio, Speech, and Language Processing}}  \bibinfo{volume}{30} (\bibinfo{year}{2022}), \bibinfo{pages}{1386--1395}.
\newblock


\bibitem[Kingma(2014)]%
        {kingma2014adam}
\bibfield{author}{\bibinfo{person}{Diederik~P Kingma}.} \bibinfo{year}{2014}\natexlab{}.
\newblock \showarticletitle{Adam: A method for stochastic optimization}.
\newblock \bibinfo{journal}{\emph{arXiv preprint arXiv:1412.6980}} (\bibinfo{year}{2014}).
\newblock


\bibitem[Krause et~al\mbox{.}(2013)]%
        {krause20133d}
\bibfield{author}{\bibinfo{person}{Jonathan Krause}, \bibinfo{person}{Michael Stark}, \bibinfo{person}{Jia Deng}, {and} \bibinfo{person}{Li Fei-Fei}.} \bibinfo{year}{2013}\natexlab{}.
\newblock \showarticletitle{3d object representations for fine-grained categorization}. In \bibinfo{booktitle}{\emph{Proceedings of the IEEE international conference on computer vision workshops}}. \bibinfo{pages}{554--561}.
\newblock


\bibitem[Krizhevsky et~al\mbox{.}(2009)]%
        {krizhevsky2009learning}
\bibfield{author}{\bibinfo{person}{Alex Krizhevsky}, \bibinfo{person}{Geoffrey Hinton}, {et~al\mbox{.}}} \bibinfo{year}{2009}\natexlab{}.
\newblock \showarticletitle{Learning multiple layers of features from tiny images}.
\newblock  (\bibinfo{year}{2009}).
\newblock


\bibitem[Lee et~al\mbox{.}(2018)]%
        {lee2018simple}
\bibfield{author}{\bibinfo{person}{Kimin Lee}, \bibinfo{person}{Kibok Lee}, \bibinfo{person}{Honglak Lee}, {and} \bibinfo{person}{Jinwoo Shin}.} \bibinfo{year}{2018}\natexlab{}.
\newblock \showarticletitle{A simple unified framework for detecting out-of-distribution samples and adversarial attacks}.
\newblock \bibinfo{journal}{\emph{Advances in neural information processing systems}}  \bibinfo{volume}{31} (\bibinfo{year}{2018}).
\newblock


\bibitem[Li et~al\mbox{.}(2024)]%
        {li2024learning}
\bibfield{author}{\bibinfo{person}{Tianqi Li}, \bibinfo{person}{Guansong Pang}, \bibinfo{person}{Xiao Bai}, \bibinfo{person}{Wenjun Miao}, {and} \bibinfo{person}{Jin Zheng}.} \bibinfo{year}{2024}\natexlab{}.
\newblock \showarticletitle{Learning transferable negative prompts for out-of-distribution detection}. In \bibinfo{booktitle}{\emph{Proceedings of the IEEE/CVF Conference on Computer Vision and Pattern Recognition}}. \bibinfo{pages}{17584--17594}.
\newblock


\bibitem[Lian et~al\mbox{.}(2022)]%
        {lian2022scaling}
\bibfield{author}{\bibinfo{person}{Dongze Lian}, \bibinfo{person}{Daquan Zhou}, \bibinfo{person}{Jiashi Feng}, {and} \bibinfo{person}{Xinchao Wang}.} \bibinfo{year}{2022}\natexlab{}.
\newblock \showarticletitle{Scaling \& shifting your features: A new baseline for efficient model tuning}.
\newblock \bibinfo{journal}{\emph{Advances in Neural Information Processing Systems}}  \bibinfo{volume}{35} (\bibinfo{year}{2022}), \bibinfo{pages}{109--123}.
\newblock


\bibitem[Liang et~al\mbox{.}(2017)]%
        {liang2017enhancing}
\bibfield{author}{\bibinfo{person}{Shiyu Liang}, \bibinfo{person}{Yixuan Li}, {and} \bibinfo{person}{Rayadurgam Srikant}.} \bibinfo{year}{2017}\natexlab{}.
\newblock \showarticletitle{Enhancing the reliability of out-of-distribution image detection in neural networks}.
\newblock \bibinfo{journal}{\emph{arXiv preprint arXiv:1706.02690}} (\bibinfo{year}{2017}).
\newblock


\bibitem[Lin et~al\mbox{.}(2021)]%
        {lin2021mood}
\bibfield{author}{\bibinfo{person}{Ziqian Lin}, \bibinfo{person}{Sreya~Dutta Roy}, {and} \bibinfo{person}{Yixuan Li}.} \bibinfo{year}{2021}\natexlab{}.
\newblock \showarticletitle{Mood: Multi-level out-of-distribution detection}. In \bibinfo{booktitle}{\emph{Proceedings of the IEEE/CVF conference on Computer Vision and Pattern Recognition}}. \bibinfo{pages}{15313--15323}.
\newblock


\bibitem[Liu et~al\mbox{.}(2020)]%
        {liu2020energy}
\bibfield{author}{\bibinfo{person}{Weitang Liu}, \bibinfo{person}{Xiaoyun Wang}, \bibinfo{person}{John Owens}, {and} \bibinfo{person}{Yixuan Li}.} \bibinfo{year}{2020}\natexlab{}.
\newblock \showarticletitle{Energy-based out-of-distribution detection}.
\newblock \bibinfo{journal}{\emph{Advances in neural information processing systems}}  \bibinfo{volume}{33} (\bibinfo{year}{2020}), \bibinfo{pages}{21464--21475}.
\newblock


\bibitem[Liu et~al\mbox{.}(2023)]%
        {liu2023gen}
\bibfield{author}{\bibinfo{person}{Xixi Liu}, \bibinfo{person}{Yaroslava Lochman}, {and} \bibinfo{person}{Christopher Zach}.} \bibinfo{year}{2023}\natexlab{}.
\newblock \showarticletitle{Gen: Pushing the limits of softmax-based out-of-distribution detection}. In \bibinfo{booktitle}{\emph{Proceedings of the IEEE/CVF conference on computer vision and pattern recognition}}. \bibinfo{pages}{23946--23955}.
\newblock


\bibitem[Ming et~al\mbox{.}(2022a)]%
        {ming2022delving}
\bibfield{author}{\bibinfo{person}{Yifei Ming}, \bibinfo{person}{Ziyang Cai}, \bibinfo{person}{Jiuxiang Gu}, \bibinfo{person}{Yiyou Sun}, \bibinfo{person}{Wei Li}, {and} \bibinfo{person}{Yixuan Li}.} \bibinfo{year}{2022}\natexlab{a}.
\newblock \showarticletitle{Delving into out-of-distribution detection with vision-language representations}.
\newblock \bibinfo{journal}{\emph{Advances in neural information processing systems}}  \bibinfo{volume}{35} (\bibinfo{year}{2022}), \bibinfo{pages}{35087--35102}.
\newblock


\bibitem[Ming et~al\mbox{.}(2022b)]%
        {ming2022poem}
\bibfield{author}{\bibinfo{person}{Yifei Ming}, \bibinfo{person}{Ying Fan}, {and} \bibinfo{person}{Yixuan Li}.} \bibinfo{year}{2022}\natexlab{b}.
\newblock \showarticletitle{Poem: Out-of-distribution detection with posterior sampling}. In \bibinfo{booktitle}{\emph{International Conference on Machine Learning}}. PMLR, \bibinfo{pages}{15650--15665}.
\newblock


\bibitem[Ming et~al\mbox{.}(2022c)]%
        {ming2022exploit}
\bibfield{author}{\bibinfo{person}{Yifei Ming}, \bibinfo{person}{Yiyou Sun}, \bibinfo{person}{Ousmane Dia}, {and} \bibinfo{person}{Yixuan Li}.} \bibinfo{year}{2022}\natexlab{c}.
\newblock \showarticletitle{How to exploit hyperspherical embeddings for out-of-distribution detection?}
\newblock \bibinfo{journal}{\emph{arXiv preprint arXiv:2203.04450}} (\bibinfo{year}{2022}).
\newblock


\bibitem[Miyai et~al\mbox{.}(2024)]%
        {miyai2024locoop}
\bibfield{author}{\bibinfo{person}{Atsuyuki Miyai}, \bibinfo{person}{Qing Yu}, \bibinfo{person}{Go Irie}, {and} \bibinfo{person}{Kiyoharu Aizawa}.} \bibinfo{year}{2024}\natexlab{}.
\newblock \showarticletitle{Locoop: Few-shot out-of-distribution detection via prompt learning}.
\newblock \bibinfo{journal}{\emph{Advances in Neural Information Processing Systems}}  \bibinfo{volume}{36} (\bibinfo{year}{2024}).
\newblock


\bibitem[Netzer et~al\mbox{.}(2011)]%
        {netzer2011reading}
\bibfield{author}{\bibinfo{person}{Yuval Netzer}, \bibinfo{person}{Tao Wang}, \bibinfo{person}{Adam Coates}, \bibinfo{person}{Alessandro Bissacco}, \bibinfo{person}{Baolin Wu}, \bibinfo{person}{Andrew~Y Ng}, {et~al\mbox{.}}} \bibinfo{year}{2011}\natexlab{}.
\newblock \showarticletitle{Reading digits in natural images with unsupervised feature learning}. In \bibinfo{booktitle}{\emph{NIPS workshop on deep learning and unsupervised feature learning}}, Vol.~\bibinfo{volume}{2011}. Granada, \bibinfo{pages}{4}.
\newblock


\bibitem[Nguyen et~al\mbox{.}(2015)]%
        {nguyen2015deep}
\bibfield{author}{\bibinfo{person}{Anh Nguyen}, \bibinfo{person}{Jason Yosinski}, {and} \bibinfo{person}{Jeff Clune}.} \bibinfo{year}{2015}\natexlab{}.
\newblock \showarticletitle{Deep neural networks are easily fooled: High confidence predictions for unrecognizable images}. In \bibinfo{booktitle}{\emph{Proceedings of the IEEE conference on computer vision and pattern recognition}}. \bibinfo{pages}{427--436}.
\newblock


\bibitem[Nie et~al\mbox{.}(2024)]%
        {nie2024out}
\bibfield{author}{\bibinfo{person}{Jun Nie}, \bibinfo{person}{Yonggang Zhang}, \bibinfo{person}{Zhen Fang}, \bibinfo{person}{Tongliang Liu}, \bibinfo{person}{Bo Han}, {and} \bibinfo{person}{Xinmei Tian}.} \bibinfo{year}{2024}\natexlab{}.
\newblock \showarticletitle{Out-of-Distribution Detection with Negative Prompts}. In \bibinfo{booktitle}{\emph{The Twelfth International Conference on Learning Representations}}.
\newblock


\bibitem[Papadopoulos et~al\mbox{.}(2021)]%
        {papadopoulos2021outlier}
\bibfield{author}{\bibinfo{person}{Aristotelis-Angelos Papadopoulos}, \bibinfo{person}{Mohammad~Reza Rajati}, \bibinfo{person}{Nazim Shaikh}, {and} \bibinfo{person}{Jiamian Wang}.} \bibinfo{year}{2021}\natexlab{}.
\newblock \showarticletitle{Outlier exposure with confidence control for out-of-distribution detection}.
\newblock \bibinfo{journal}{\emph{Neurocomputing}}  \bibinfo{volume}{441} (\bibinfo{year}{2021}), \bibinfo{pages}{138--150}.
\newblock


\bibitem[Park et~al\mbox{.}(2023)]%
        {park2023nearest}
\bibfield{author}{\bibinfo{person}{Jaewoo Park}, \bibinfo{person}{Yoon~Gyo Jung}, {and} \bibinfo{person}{Andrew Beng~Jin Teoh}.} \bibinfo{year}{2023}\natexlab{}.
\newblock \showarticletitle{Nearest neighbor guidance for out-of-distribution detection}. In \bibinfo{booktitle}{\emph{Proceedings of the IEEE/CVF International Conference on Computer Vision}}. \bibinfo{pages}{1686--1695}.
\newblock


\bibitem[Parkhi et~al\mbox{.}(2012)]%
        {parkhi2012cats}
\bibfield{author}{\bibinfo{person}{Omkar~M Parkhi}, \bibinfo{person}{Andrea Vedaldi}, \bibinfo{person}{Andrew Zisserman}, {and} \bibinfo{person}{CV Jawahar}.} \bibinfo{year}{2012}\natexlab{}.
\newblock \showarticletitle{Cats and dogs}. In \bibinfo{booktitle}{\emph{2012 IEEE conference on computer vision and pattern recognition}}. IEEE, \bibinfo{pages}{3498--3505}.
\newblock


\bibitem[Pidhorskyi et~al\mbox{.}(2018)]%
        {pidhorskyi2018generative}
\bibfield{author}{\bibinfo{person}{Stanislav Pidhorskyi}, \bibinfo{person}{Ranya Almohsen}, {and} \bibinfo{person}{Gianfranco Doretto}.} \bibinfo{year}{2018}\natexlab{}.
\newblock \showarticletitle{Generative probabilistic novelty detection with adversarial autoencoders}.
\newblock \bibinfo{journal}{\emph{Advances in neural information processing systems}}  \bibinfo{volume}{31} (\bibinfo{year}{2018}).
\newblock


\bibitem[Radford et~al\mbox{.}(2021)]%
        {radford2021learning}
\bibfield{author}{\bibinfo{person}{Alec Radford}, \bibinfo{person}{Jong~Wook Kim}, \bibinfo{person}{Chris Hallacy}, \bibinfo{person}{Aditya Ramesh}, \bibinfo{person}{Gabriel Goh}, \bibinfo{person}{Sandhini Agarwal}, \bibinfo{person}{Girish Sastry}, \bibinfo{person}{Amanda Askell}, \bibinfo{person}{Pamela Mishkin}, \bibinfo{person}{Jack Clark}, {et~al\mbox{.}}} \bibinfo{year}{2021}\natexlab{}.
\newblock \showarticletitle{Learning transferable visual models from natural language supervision}. In \bibinfo{booktitle}{\emph{International conference on machine learning}}. PMLR, \bibinfo{pages}{8748--8763}.
\newblock


\bibitem[Recht et~al\mbox{.}(2019)]%
        {recht2019imagenet}
\bibfield{author}{\bibinfo{person}{Benjamin Recht}, \bibinfo{person}{Rebecca Roelofs}, \bibinfo{person}{Ludwig Schmidt}, {and} \bibinfo{person}{Vaishaal Shankar}.} \bibinfo{year}{2019}\natexlab{}.
\newblock \showarticletitle{Do imagenet classifiers generalize to imagenet?}. In \bibinfo{booktitle}{\emph{International conference on machine learning}}. PMLR, \bibinfo{pages}{5389--5400}.
\newblock


\bibitem[Ren et~al\mbox{.}(2021)]%
        {ren2021simple}
\bibfield{author}{\bibinfo{person}{Jie Ren}, \bibinfo{person}{Stanislav Fort}, \bibinfo{person}{Jeremiah Liu}, \bibinfo{person}{Abhijit~Guha Roy}, \bibinfo{person}{Shreyas Padhy}, {and} \bibinfo{person}{Balaji Lakshminarayanan}.} \bibinfo{year}{2021}\natexlab{}.
\newblock \showarticletitle{A simple fix to mahalanobis distance for improving near-ood detection}.
\newblock \bibinfo{journal}{\emph{arXiv preprint arXiv:2106.09022}} (\bibinfo{year}{2021}).
\newblock


\bibitem[Ren et~al\mbox{.}(2019)]%
        {ren2019likelihood}
\bibfield{author}{\bibinfo{person}{Jie Ren}, \bibinfo{person}{Peter~J Liu}, \bibinfo{person}{Emily Fertig}, \bibinfo{person}{Jasper Snoek}, \bibinfo{person}{Ryan Poplin}, \bibinfo{person}{Mark Depristo}, \bibinfo{person}{Joshua Dillon}, {and} \bibinfo{person}{Balaji Lakshminarayanan}.} \bibinfo{year}{2019}\natexlab{}.
\newblock \showarticletitle{Likelihood ratios for out-of-distribution detection}.
\newblock \bibinfo{journal}{\emph{Advances in neural information processing systems}}  \bibinfo{volume}{32} (\bibinfo{year}{2019}).
\newblock


\bibitem[Sastry and Oore(2020)]%
        {sastry2020detecting}
\bibfield{author}{\bibinfo{person}{Chandramouli~Shama Sastry} {and} \bibinfo{person}{Sageev Oore}.} \bibinfo{year}{2020}\natexlab{}.
\newblock \showarticletitle{Detecting out-of-distribution examples with gram matrices}. In \bibinfo{booktitle}{\emph{International Conference on Machine Learning}}. PMLR, \bibinfo{pages}{8491--8501}.
\newblock


\bibitem[Scheirer et~al\mbox{.}(2012)]%
        {scheirer2012toward}
\bibfield{author}{\bibinfo{person}{Walter~J Scheirer}, \bibinfo{person}{Anderson de Rezende~Rocha}, \bibinfo{person}{Archana Sapkota}, {and} \bibinfo{person}{Terrance~E Boult}.} \bibinfo{year}{2012}\natexlab{}.
\newblock \showarticletitle{Toward open set recognition}.
\newblock \bibinfo{journal}{\emph{IEEE transactions on pattern analysis and machine intelligence}} \bibinfo{volume}{35}, \bibinfo{number}{7} (\bibinfo{year}{2012}), \bibinfo{pages}{1757--1772}.
\newblock


\bibitem[Sun et~al\mbox{.}(2021)]%
        {sun2021react}
\bibfield{author}{\bibinfo{person}{Yiyou Sun}, \bibinfo{person}{Chuan Guo}, {and} \bibinfo{person}{Yixuan Li}.} \bibinfo{year}{2021}\natexlab{}.
\newblock \showarticletitle{React: Out-of-distribution detection with rectified activations}.
\newblock \bibinfo{journal}{\emph{Advances in Neural Information Processing Systems}}  \bibinfo{volume}{34} (\bibinfo{year}{2021}), \bibinfo{pages}{144--157}.
\newblock


\bibitem[Sun and Li(2022)]%
        {sun2022dice}
\bibfield{author}{\bibinfo{person}{Yiyou Sun} {and} \bibinfo{person}{Yixuan Li}.} \bibinfo{year}{2022}\natexlab{}.
\newblock \showarticletitle{Dice: Leveraging sparsification for out-of-distribution detection}. In \bibinfo{booktitle}{\emph{European Conference on Computer Vision}}. Springer, \bibinfo{pages}{691--708}.
\newblock


\bibitem[Sun et~al\mbox{.}(2022)]%
        {sun2022out}
\bibfield{author}{\bibinfo{person}{Yiyou Sun}, \bibinfo{person}{Yifei Ming}, \bibinfo{person}{Xiaojin Zhu}, {and} \bibinfo{person}{Yixuan Li}.} \bibinfo{year}{2022}\natexlab{}.
\newblock \showarticletitle{Out-of-distribution detection with deep nearest neighbors}. In \bibinfo{booktitle}{\emph{International Conference on Machine Learning}}. PMLR, \bibinfo{pages}{20827--20840}.
\newblock


\bibitem[Tao et~al\mbox{.}(2023)]%
        {tao2023non}
\bibfield{author}{\bibinfo{person}{Leitian Tao}, \bibinfo{person}{Xuefeng Du}, \bibinfo{person}{Xiaojin Zhu}, {and} \bibinfo{person}{Yixuan Li}.} \bibinfo{year}{2023}\natexlab{}.
\newblock \showarticletitle{Non-parametric outlier synthesis}.
\newblock \bibinfo{journal}{\emph{arXiv preprint arXiv:2303.02966}} (\bibinfo{year}{2023}).
\newblock


\bibitem[Thulasidasan et~al\mbox{.}(2019)]%
        {thulasidasan2019mixup}
\bibfield{author}{\bibinfo{person}{Sunil Thulasidasan}, \bibinfo{person}{Gopinath Chennupati}, \bibinfo{person}{Jeff~A Bilmes}, \bibinfo{person}{Tanmoy Bhattacharya}, {and} \bibinfo{person}{Sarah Michalak}.} \bibinfo{year}{2019}\natexlab{}.
\newblock \showarticletitle{On mixup training: Improved calibration and predictive uncertainty for deep neural networks}.
\newblock \bibinfo{journal}{\emph{Advances in neural information processing systems}}  \bibinfo{volume}{32} (\bibinfo{year}{2019}).
\newblock


\bibitem[Van~Amersfoort et~al\mbox{.}(2020)]%
        {van2020uncertainty}
\bibfield{author}{\bibinfo{person}{Joost Van~Amersfoort}, \bibinfo{person}{Lewis Smith}, \bibinfo{person}{Yee~Whye Teh}, {and} \bibinfo{person}{Yarin Gal}.} \bibinfo{year}{2020}\natexlab{}.
\newblock \showarticletitle{Uncertainty estimation using a single deep deterministic neural network}. In \bibinfo{booktitle}{\emph{International conference on machine learning}}. PMLR, \bibinfo{pages}{9690--9700}.
\newblock


\bibitem[Van~Horn et~al\mbox{.}(2018)]%
        {van2018inaturalist}
\bibfield{author}{\bibinfo{person}{Grant Van~Horn}, \bibinfo{person}{Oisin Mac~Aodha}, \bibinfo{person}{Yang Song}, \bibinfo{person}{Yin Cui}, \bibinfo{person}{Chen Sun}, \bibinfo{person}{Alex Shepard}, \bibinfo{person}{Hartwig Adam}, \bibinfo{person}{Pietro Perona}, {and} \bibinfo{person}{Serge Belongie}.} \bibinfo{year}{2018}\natexlab{}.
\newblock \showarticletitle{The inaturalist species classification and detection dataset}. In \bibinfo{booktitle}{\emph{Proceedings of the IEEE conference on computer vision and pattern recognition}}. \bibinfo{pages}{8769--8778}.
\newblock


\bibitem[Wah et~al\mbox{.}(2011)]%
        {wah2011caltech}
\bibfield{author}{\bibinfo{person}{Catherine Wah}, \bibinfo{person}{Steve Branson}, \bibinfo{person}{Peter Welinder}, \bibinfo{person}{Pietro Perona}, {and} \bibinfo{person}{Serge Belongie}.} \bibinfo{year}{2011}\natexlab{}.
\newblock \showarticletitle{The caltech-ucsd birds-200-2011 dataset}.
\newblock  (\bibinfo{year}{2011}).
\newblock


\bibitem[Wang et~al\mbox{.}(2019)]%
        {wang2019learning}
\bibfield{author}{\bibinfo{person}{Haohan Wang}, \bibinfo{person}{Songwei Ge}, \bibinfo{person}{Zachary Lipton}, {and} \bibinfo{person}{Eric~P Xing}.} \bibinfo{year}{2019}\natexlab{}.
\newblock \showarticletitle{Learning Robust Global Representations by Penalizing Local Predictive Power}. In \bibinfo{booktitle}{\emph{Advances in Neural Information Processing Systems}}. \bibinfo{pages}{10506--10518}.
\newblock


\bibitem[Wang et~al\mbox{.}(2023)]%
        {wang2023clipn}
\bibfield{author}{\bibinfo{person}{Hualiang Wang}, \bibinfo{person}{Yi Li}, \bibinfo{person}{Huifeng Yao}, {and} \bibinfo{person}{Xiaomeng Li}.} \bibinfo{year}{2023}\natexlab{}.
\newblock \showarticletitle{Clipn for zero-shot ood detection: Teaching clip to say no}. In \bibinfo{booktitle}{\emph{Proceedings of the IEEE/CVF International Conference on Computer Vision}}. \bibinfo{pages}{1802--1812}.
\newblock


\bibitem[Wang et~al\mbox{.}(2022)]%
        {wang2022vim}
\bibfield{author}{\bibinfo{person}{Haoqi Wang}, \bibinfo{person}{Zhizhong Li}, \bibinfo{person}{Litong Feng}, {and} \bibinfo{person}{Wayne Zhang}.} \bibinfo{year}{2022}\natexlab{}.
\newblock \showarticletitle{Vim: Out-of-distribution with virtual-logit matching}. In \bibinfo{booktitle}{\emph{Proceedings of the IEEE/CVF conference on computer vision and pattern recognition}}. \bibinfo{pages}{4921--4930}.
\newblock


\bibitem[Wei et~al\mbox{.}(2022)]%
        {wei2022mitigating}
\bibfield{author}{\bibinfo{person}{Hongxin Wei}, \bibinfo{person}{Renchunzi Xie}, \bibinfo{person}{Hao Cheng}, \bibinfo{person}{Lei Feng}, \bibinfo{person}{Bo An}, {and} \bibinfo{person}{Yixuan Li}.} \bibinfo{year}{2022}\natexlab{}.
\newblock \showarticletitle{Mitigating neural network overconfidence with logit normalization}. In \bibinfo{booktitle}{\emph{International conference on machine learning}}. PMLR, \bibinfo{pages}{23631--23644}.
\newblock


\bibitem[Xiao et~al\mbox{.}(2010)]%
        {xiao2010sun}
\bibfield{author}{\bibinfo{person}{Jianxiong Xiao}, \bibinfo{person}{James Hays}, \bibinfo{person}{Krista~A Ehinger}, \bibinfo{person}{Aude Oliva}, {and} \bibinfo{person}{Antonio Torralba}.} \bibinfo{year}{2010}\natexlab{}.
\newblock \showarticletitle{Sun database: Large-scale scene recognition from abbey to zoo}. In \bibinfo{booktitle}{\emph{2010 IEEE computer society conference on computer vision and pattern recognition}}. IEEE, \bibinfo{pages}{3485--3492}.
\newblock


\bibitem[Yang et~al\mbox{.}(2022)]%
        {yang2022openood}
\bibfield{author}{\bibinfo{person}{Jingkang Yang}, \bibinfo{person}{Pengyun Wang}, \bibinfo{person}{Dejian Zou}, \bibinfo{person}{Zitang Zhou}, \bibinfo{person}{Kunyuan Ding}, \bibinfo{person}{Wenxuan Peng}, \bibinfo{person}{Haoqi Wang}, \bibinfo{person}{Guangyao Chen}, \bibinfo{person}{Bo Li}, \bibinfo{person}{Yiyou Sun}, {et~al\mbox{.}}} \bibinfo{year}{2022}\natexlab{}.
\newblock \showarticletitle{Openood: Benchmarking generalized out-of-distribution detection}.
\newblock \bibinfo{journal}{\emph{Advances in Neural Information Processing Systems}}  \bibinfo{volume}{35} (\bibinfo{year}{2022}), \bibinfo{pages}{32598--32611}.
\newblock


\bibitem[Yu et~al\mbox{.}(2024)]%
        {yu2024self}
\bibfield{author}{\bibinfo{person}{Geng Yu}, \bibinfo{person}{Jianing Zhu}, \bibinfo{person}{Jiangchao Yao}, {and} \bibinfo{person}{Bo Han}.} \bibinfo{year}{2024}\natexlab{}.
\newblock \showarticletitle{Self-Calibrated Tuning of Vision-Language Models for Out-of-Distribution Detection}.
\newblock \bibinfo{journal}{\emph{Advances in Neural Information Processing Systems}}  \bibinfo{volume}{37} (\bibinfo{year}{2024}), \bibinfo{pages}{56322--56348}.
\newblock


\bibitem[Yu and Aizawa(2019)]%
        {yu2019unsupervised}
\bibfield{author}{\bibinfo{person}{Qing Yu} {and} \bibinfo{person}{Kiyoharu Aizawa}.} \bibinfo{year}{2019}\natexlab{}.
\newblock \showarticletitle{Unsupervised out-of-distribution detection by maximum classifier discrepancy}. In \bibinfo{booktitle}{\emph{Proceedings of the IEEE/CVF international conference on computer vision}}. \bibinfo{pages}{9518--9526}.
\newblock


\bibitem[Yu et~al\mbox{.}(2023)]%
        {yu2023task}
\bibfield{author}{\bibinfo{person}{Tao Yu}, \bibinfo{person}{Zhihe Lu}, \bibinfo{person}{Xin Jin}, \bibinfo{person}{Zhibo Chen}, {and} \bibinfo{person}{Xinchao Wang}.} \bibinfo{year}{2023}\natexlab{}.
\newblock \showarticletitle{Task residual for tuning vision-language models}. In \bibinfo{booktitle}{\emph{Proceedings of the IEEE/CVF Conference on Computer Vision and Pattern Recognition}}. \bibinfo{pages}{10899--10909}.
\newblock


\bibitem[Yun et~al\mbox{.}(2019)]%
        {yun2019cutmix}
\bibfield{author}{\bibinfo{person}{Sangdoo Yun}, \bibinfo{person}{Dongyoon Han}, \bibinfo{person}{Seong~Joon Oh}, \bibinfo{person}{Sanghyuk Chun}, \bibinfo{person}{Junsuk Choe}, {and} \bibinfo{person}{Youngjoon Yoo}.} \bibinfo{year}{2019}\natexlab{}.
\newblock \showarticletitle{Cutmix: Regularization strategy to train strong classifiers with localizable features}. In \bibinfo{booktitle}{\emph{Proceedings of the IEEE/CVF international conference on computer vision}}. \bibinfo{pages}{6023--6032}.
\newblock


\bibitem[Zaeemzadeh et~al\mbox{.}(2021)]%
        {zaeemzadeh2021out}
\bibfield{author}{\bibinfo{person}{Alireza Zaeemzadeh}, \bibinfo{person}{Niccolo Bisagno}, \bibinfo{person}{Zeno Sambugaro}, \bibinfo{person}{Nicola Conci}, \bibinfo{person}{Nazanin Rahnavard}, {and} \bibinfo{person}{Mubarak Shah}.} \bibinfo{year}{2021}\natexlab{}.
\newblock \showarticletitle{Out-of-distribution detection using union of 1-dimensional subspaces}. In \bibinfo{booktitle}{\emph{Proceedings of the IEEE/CVF conference on Computer Vision and Pattern Recognition}}. \bibinfo{pages}{9452--9461}.
\newblock


\bibitem[Zanella and Ben~Ayed(2024)]%
        {zanella2024low}
\bibfield{author}{\bibinfo{person}{Maxime Zanella} {and} \bibinfo{person}{Ismail Ben~Ayed}.} \bibinfo{year}{2024}\natexlab{}.
\newblock \showarticletitle{Low-rank few-shot adaptation of vision-language models}. In \bibinfo{booktitle}{\emph{Proceedings of the IEEE/CVF Conference on Computer Vision and Pattern Recognition}}. \bibinfo{pages}{1593--1603}.
\newblock


\bibitem[Zhang et~al\mbox{.}(2022)]%
        {zhang2022out}
\bibfield{author}{\bibinfo{person}{Jinsong Zhang}, \bibinfo{person}{Qiang Fu}, \bibinfo{person}{Xu Chen}, \bibinfo{person}{Lun Du}, \bibinfo{person}{Zelin Li}, \bibinfo{person}{Gang Wang}, \bibinfo{person}{Shi Han}, \bibinfo{person}{Dongmei Zhang}, {et~al\mbox{.}}} \bibinfo{year}{2022}\natexlab{}.
\newblock \showarticletitle{Out-of-distribution detection based on in-distribution data patterns memorization with modern hopfield energy}. In \bibinfo{booktitle}{\emph{The Eleventh International Conference on Learning Representations}}.
\newblock


\bibitem[Zhang and Zhang(2024)]%
        {zhang2024adaneg}
\bibfield{author}{\bibinfo{person}{Yabin Zhang} {and} \bibinfo{person}{Lei Zhang}.} \bibinfo{year}{2024}\natexlab{}.
\newblock \showarticletitle{Adaneg: Adaptive negative proxy guided ood detection with vision-language models}.
\newblock \bibinfo{journal}{\emph{Advances in Neural Information Processing Systems}}  \bibinfo{volume}{37} (\bibinfo{year}{2024}), \bibinfo{pages}{38744--38768}.
\newblock


\bibitem[Zhang et~al\mbox{.}(2024a)]%
        {zhang2024lapt}
\bibfield{author}{\bibinfo{person}{Yabin Zhang}, \bibinfo{person}{Wenjie Zhu}, \bibinfo{person}{Chenhang He}, {and} \bibinfo{person}{Lei Zhang}.} \bibinfo{year}{2024}\natexlab{a}.
\newblock \showarticletitle{Lapt: Label-driven automated prompt tuning for ood detection with vision-language models}.
\newblock \bibinfo{journal}{\emph{arXiv preprint arXiv:2407.08966}} (\bibinfo{year}{2024}).
\newblock


\bibitem[Zhang et~al\mbox{.}(2024b)]%
        {zhang2024dual}
\bibfield{author}{\bibinfo{person}{Yabin Zhang}, \bibinfo{person}{Wenjie Zhu}, \bibinfo{person}{Hui Tang}, \bibinfo{person}{Zhiyuan Ma}, \bibinfo{person}{Kaiyang Zhou}, {and} \bibinfo{person}{Lei Zhang}.} \bibinfo{year}{2024}\natexlab{b}.
\newblock \showarticletitle{Dual memory networks: A versatile adaptation approach for vision-language models}. In \bibinfo{booktitle}{\emph{Proceedings of the IEEE/CVF conference on computer vision and pattern recognition}}. \bibinfo{pages}{28718--28728}.
\newblock


\bibitem[Zhou et~al\mbox{.}(2017)]%
        {zhou2017places}
\bibfield{author}{\bibinfo{person}{Bolei Zhou}, \bibinfo{person}{Agata Lapedriza}, \bibinfo{person}{Aditya Khosla}, \bibinfo{person}{Aude Oliva}, {and} \bibinfo{person}{Antonio Torralba}.} \bibinfo{year}{2017}\natexlab{}.
\newblock \showarticletitle{Places: A 10 million image database for scene recognition}.
\newblock \bibinfo{journal}{\emph{IEEE transactions on pattern analysis and machine intelligence}} \bibinfo{volume}{40}, \bibinfo{number}{6} (\bibinfo{year}{2017}), \bibinfo{pages}{1452--1464}.
\newblock


\bibitem[Zhou et~al\mbox{.}(2022a)]%
        {zhou2022conditional}
\bibfield{author}{\bibinfo{person}{Kaiyang Zhou}, \bibinfo{person}{Jingkang Yang}, \bibinfo{person}{Chen~Change Loy}, {and} \bibinfo{person}{Ziwei Liu}.} \bibinfo{year}{2022}\natexlab{a}.
\newblock \showarticletitle{Conditional prompt learning for vision-language models}. In \bibinfo{booktitle}{\emph{Proceedings of the IEEE/CVF conference on computer vision and pattern recognition}}. \bibinfo{pages}{16816--16825}.
\newblock


\bibitem[Zhou et~al\mbox{.}(2022b)]%
        {zhou2022learning}
\bibfield{author}{\bibinfo{person}{Kaiyang Zhou}, \bibinfo{person}{Jingkang Yang}, \bibinfo{person}{Chen~Change Loy}, {and} \bibinfo{person}{Ziwei Liu}.} \bibinfo{year}{2022}\natexlab{b}.
\newblock \showarticletitle{Learning to prompt for vision-language models}.
\newblock \bibinfo{journal}{\emph{International Journal of Computer Vision}} \bibinfo{volume}{130}, \bibinfo{number}{9} (\bibinfo{year}{2022}), \bibinfo{pages}{2337--2348}.
\newblock


\bibitem[Zong et~al\mbox{.}(2018)]%
        {zong2018deep}
\bibfield{author}{\bibinfo{person}{Bo Zong}, \bibinfo{person}{Qi Song}, \bibinfo{person}{Martin~Renqiang Min}, \bibinfo{person}{Wei Cheng}, \bibinfo{person}{Cristian Lumezanu}, \bibinfo{person}{Daeki Cho}, {and} \bibinfo{person}{Haifeng Chen}.} \bibinfo{year}{2018}\natexlab{}.
\newblock \showarticletitle{Deep autoencoding gaussian mixture model for unsupervised anomaly detection}. In \bibinfo{booktitle}{\emph{International conference on learning representations}}.
\newblock


\end{thebibliography}

\appendix
\end{document}